\documentclass[sigconf,nonacm,balance=false]{acmart}

\usepackage{xspace}
\usepackage{amsmath,amsfonts,mathtools}
\usepackage{booktabs}
\usepackage{multirow}
\usepackage{tabularx}
\usepackage{algorithm}
\usepackage{algorithmic}
\usepackage{enumitem}
\usepackage{array}
\usepackage{graphicx}
\usepackage{microtype}
\usepackage{url}
\usepackage{tikz}
\usetikzlibrary{arrows.meta,positioning,calc}

\newcommand{\fm}{\textsc{FM4NILM}\xspace}

\newcommand{\softplus}{\operatorname{softplus}}
\newcommand{\sigmoid}{\operatorname{sigmoid}}

\AtBeginDocument{%
  \setlength{\abovecaptionskip}{5pt}%
  \setlength{\belowcaptionskip}{4pt}%
  \setlength{\textfloatsep}{10pt plus 4pt minus 3pt}%
  \setlength{\floatsep}{8pt plus 3pt minus 2pt}%
  \setlength{\intextsep}{9pt plus 3pt minus 2pt}%
  \setlength{\dbltextfloatsep}{10pt plus 4pt minus 3pt}%
  \setlength{\dblfloatsep}{8pt plus 3pt minus 2pt}%
}

\title[Ask for Any Appliance: A Prompt-Programmable Foundation Model for Non-Intrusive Load Monitoring]{Ask for Any Appliance: A Prompt-Programmable Foundation Model for Non-Intrusive Load Monitoring}

\setcopyright{none}
\acmDOI{}
\acmISBN{}
\fancypagestyle{preprint}{
  \fancyhf{}
  \fancyhead[L]{\small Preprint}
  \fancyfoot[C]{\small\thepage}

}

\author{Xudong Wang}
\affiliation{
  \institution{The Chinese University of Hong Kong, Shenzhen}
  \city{Shenzhen}
  \country{China}
}
\email{xudongwang@link.cuhk.edu.cn}

\author{Jiacheng Cui}
\affiliation{
  \institution{The Hong Kong University of Science and Technology (Guangzhou)}
  \city{Guangzhou}
  \country{China}
}
\email{jcui047@connect.hkust-gz.edu.cn}

\author{Junyu Xue}
\affiliation{
  \institution{Southern University of Science and Technology}
  \city{Shenzhen}
  \country{China}
}
\email{junyuxue@outlook.com}

\author{Tongxin Li}
\affiliation{
  \institution{The Chinese University of Hong Kong, Shenzhen}
  \city{Shenzhen}
  \country{China}
}
\email{litongxin@cuhk.edu.cn}

\author{Guoming Tang}
\authornote{Corresponding author.}
\affiliation{
  \institution{The Hong Kong University of Science and Technology (Guangzhou)}
  \city{Guangzhou}
  \country{China}
}
\email{guomingtang@hkust-gz.edu.cn}
\renewcommand{\shortauthors}{Xudong Wang et al.}

\begin{document}

\begin{abstract}

Non-intrusive load monitoring (NILM) estimates appliance-level consumption
from a single whole-home meter, but many deep learning approaches remain
tied to appliance-specific models or fixed output inventories.
This fragmentation increases deployment and maintenance costs and makes
appliance coverage difficult to extend without retraining.
We present \fm (Foundation Model for NILM), a single,
prompt-programmable model that takes appliance identity as part of its
input. Given aggregate measurements and a request comprising a
natural-language description and optional measured activation exemplars,
\fm estimates the requested appliance's power trajectory.
A lightweight cadence-aware transformer is pretrained by masked
reconstruction on 645k sequences from seven public corpora spanning
1--60\,s sampling intervals, then aligned with appliance requests using
observation-masked losses for partially labeled households.
A Bernoulli--lognormal decoder separates activity detection from
conditional power estimation.
On held-out households and time periods from REDD, UK-DALE, and REFIT,
one frozen, text-prompted \fm model serves twelve appliance--corpus
requests, achieving 0.556 event F1 and 0.625 AUPRC, while attaining the lowest active-window MAE (251.8\,W) among all seven appliance-specific baselines. Streaming score aggregation further improves event F1 to 0.582 with a 60\,s aggregation delay.
Beyond these supervised requests, a separate category-held-out
evaluation shows that adding ten activation exemplars to the text
prompt raises microwave AUPRC from 0.132 to 0.214 without
parameter updates. Complementary input-intervention ablations also explore \fm's performance under different request-based behaviors like mismatching the prompt. These results show that a single prompt-programmable model can match or exceed a fleet of appliance-specific networks while also answering
requests those networks were never trained to handle, pointing toward
flexible, zero/few-shot energy disaggregation as a practical alternative to
fragmented, single-task NILM architectures.

\end{abstract}


\maketitle
\pagestyle{preprint}
\thispagestyle{preprint}
\hypersetup{
  pdfauthor={Xudong Wang, Jiacheng Cui, Junyu Xue, Tongxin Li, Guoming Tang},
  pdfsubject={Preprint},
  pdfcreator={LaTeX preprint},
  pdfkeywords={non-intrusive load monitoring, energy disaggregation, foundation models, in-context learning, language conditioning, partial labels, self-supervised pretraining}
}

\section{Introduction}
\label{sec:introduction}

Non-intrusive load monitoring (NILM) estimates appliance-level electricity
consumption from a single whole-home meter~\cite{hart1992nonintrusive}. It
underlies energy feedback programs, demand response targeting, appliance-level
diagnostics, and electrification planning, all without instrumenting every
outlet in a home. Neural models have made disaggregation steadily more
accurate~\cite{kelly2015neural,zhang2018sequence,yue2020bert4nilm,wang2022evsense,petralia2025nilmformer,xue2025prompting,wang2026energy},
but what limits NILM in the field is no longer the accuracy of any single
model, it is the number of models. Almost every published system binds one
network to one appliance in one dataset, so answering the twelve
appliance--corpus questions in this paper's own benchmark means training,
validating, and maintaining twelve networks. Every
additional appliance category, meter generation, or newly onboarded home
enlarges that inventory, and every change to it means retraining. The
appliance a household actually cares about is reachable only if someone
trained a model for it in advance.

This paper removes appliance identity from the weights and makes it an
input. Rather than a network per appliance, we ask whether one frozen model
can extract any appliance's consumption when simply told which appliance to
extract. We formalize that instruction as a \emph{request}: a short
natural-language card summarizing the appliance's typical behavior (a
microwave's brief, high-power bursts, say) and, optionally, a handful of
measured activation exemplars from other households. Given an aggregate
window and a request, the model returns the requested trajectory; a
refrigerator request and a dishwasher request posed against the same window
are answered by the same unchanged parameters. We call this
\emph{prompt-conditioned inverse inference}. Under it, extending a NILM
service to a new appliance becomes an authoring task, writing a card or
collecting a few examples, rather than a training run.

Making such an interface work means confronting three properties of energy
data that mainstream sequence pretraining never has to face. First, public corpora
submeter only a subset of a household's loads, so supervision is a mixture
whose other components are never labeled and must be excluded from the loss
rather than treated as zero. Second, real meters sample at rates differing
by more than an order of magnitude, so a shared model needs a representation
of \emph{physical} time rather than of index position. Third,
disaggregation is two estimation problems wearing one name: whether an
appliance runs, and how many watts it draws while running. These transfer
very differently across homes, since nameplate ratings and supply conditions
move the second while leaving the first intact.

\fm answers each with one architectural commitment. A cadence-aware
transformer is pretrained to reconstruct masked spans over 645k aggregate
and submeter sequences from seven public corpora spanning 1--60\,s cadences, learning
a representation of power signals before it sees any appliance label.
Supervised alignment then couples that representation to text, exemplars, or
both under observation masks that keep unlabeled loads out of the gradient.
A request-conditioned Bernoulli--lognormal decoder factorizes the output
into activity and conditional power, so detection and watt calibration are
learned, measured, and can fail independently. The resulting 3.7M-parameter,
15\,MB \fm model replaces all twelve specialists in our benchmark and
answers a request in 5.5\,ms on four CPU threads.

Our evaluation walks outward from familiar requests to unfamiliar ones, and
what it finds is as much about the interface as the accuracy. On held-out
houses and periods from REDD, UK-DALE, and REFIT, one text-prompted
\fm model tracks the best of seven appliance-specific baselines on
detection, matches it outright under validation-calibrated thresholds, and
produces the lowest active-window watt error of any model we trained. Beyond that regime, four findings characterize what the
request buys. \emph{The model reads the request:} substituting an
electrically matched confuser card, permuting the card--appliance
assignment, or blanking the aggregate each collapses accuracy, so this is
not a routing table in disguise. \emph{Language is a better address than an
index:} learned appliance IDs cost accuracy and, more tellingly, inflate
seed-to-seed variance by an order of magnitude, because the language encoder
supplies a shared geometry a free embedding must rediscover.
\emph{Exemplars pay in proportion to how much of an appliance fits inside
the prompt:} ten exemplars make a microwave excluded from every stage of
training measurable with no parameter update, yet the same budget does
nothing for a dishwasher, whose hour-long cycle no snippet can
summarize, a limit we confirm by widening the request span instead.
\emph{Pattern transfers; level does not:} a diagnostic per-household
multiplier attributes a quarter of the remaining out-of-domain watt error to
pure scale, with essentially every held-out group under-predicted. That last
finding is this interface's honest boundary, and \S\ref{sec:discussion} says
what would close it.

This paper makes three contributions:
\begin{enumerate}[leftmargin=*,itemsep=2pt,topsep=3pt]
    \item \textbf{A request interface for foundation NILM.} We recast
    disaggregation as prompt-conditioned component extraction, composing
    language cards with signal exemplars so that one frozen,
    meter-deployable \fm model answers familiar, unseen-appliance, and
    unseen-corpus requests through its input rather than its weights.
    \item \textbf{A recipe for learning from heterogeneous, partially
    labeled energy data.} Cadence-aware tokenization over physical time,
    span-masked pretraining, episodic prompt alignment, and
    observation-masked losses let a single model absorb seven corpora at the
    1--60\,s cadences found across real AMI deployments, conditions under
    which a conventional per-appliance head cannot be defined at all.
    \item \textbf{An evaluation that separates what transfers from what does
    not.} Five-seed comparisons, causal prompt interventions, appliance and
    corpus exclusion, and a power-error decomposition delineate the reach of
    the request interface, price it against a fleet of specialists in
    parameters and latency, and identify household watt calibration, not
    detection, as the principal remaining obstacle to prompt-programmable
    NILM.
\end{enumerate}
\section{Related Work}
\label{sec:related}

\paragraph{Energy disaggregation.}
NILM has progressed from combinatorial optimization and factorial hidden Markov models (FHMMs)~\cite{hart1992nonintrusive,kolter2012fhmm} to recurrent networks~\cite{kelly2015neural}, sequence-to-point convolution~\cite{zhang2018sequence}, gated objectives~\cite{shin2019subtask}, and transformer encoders~\cite{yue2020bert4nilm,yue2022eltransformer,petralia2025nilmformer}. Multi-appliance models amortize a shared trunk over a fixed bank of output heads~\cite{he2023msdc,xiong2023matnilm}, and transfer studies document substantial household and corpus shift~\cite{murray2017refit,ding2024appliancefilter}. Across this progression the serving interface is invariant: the set of answerable appliances is fixed when the weights are. \fm makes that set an argument of the forward pass instead.

\paragraph{Time-series foundation models.}
Chronos, TimesFM, Moirai, MOMENT, Timer, and Time-MoE establish that large-scale pretraining transfers across forecasting and reconstruction~\cite{ansari2025chronos2,das2024decoder,liu2025moiraimoe,goswami2024moment,liu2024timer,shi2024timemoe}, and large language model (LLM) variants attach textual descriptions to a series~\cite{jin2023timellm,gruver2023llmtime,xue2025prompting}. All of them predict a series that is observed. Disaggregation asks for one that is not: a latent component of a mixture whose remaining components are never labeled. That is why observation masking and an activity/scale factorization are load-bearing here and absent there.

\paragraph{Prompt-conditioned source extraction.}
Target-sound systems condition separation on class embeddings, enrollment audio, natural language, or joint audio--text prompts~\cite{delcroix2022soundbeam,liu2023audiosep,hai2025flexsed,cai2025detectanysound,radford2021clip}, and offer the closest structural analogy to our setting. Power differs from audio in ways that reshape the problem rather than rescale it: signals are nonnegative and carry absolute physical units a listener never needs, activations are sparse and heavy-tailed, cadence varies by an order of magnitude across the deployed meter fleet, and component labels are systematically incomplete. \fm adapts prompt-conditioned extraction to these constraints.

\paragraph{Language and adaptation in NILM}
Recent work guides disaggregation with LLMs, text embeddings, retrieval, and typed observation programs~\cite{garciamarrero2026refquery,chen2026sceneaware,liu2026lab}, collectively suggesting that language routes a request well but cannot, on its own, resolve electrically ambiguous categories. A separate line adapts parameters to a target home through meta-learning or test-time updates~\cite{wang2021tent,toirov2025continual,li2024tlinet}. \fm joins the two without touching the weights: language names the category, exemplars supply the measured behavior that language underdetermines, and adaptation happens entirely in the request.
\section{Problem Formulation}
\label{sec:problem}

\paragraph{Partially labeled mixtures.}
A household $h$ produces aggregate power
\begin{equation}
  x_{h,t}=\sum_{a\in\mathcal{A}_h}y_{h,a,t}+\varepsilon_{h,t},
  \label{eq:mixture}
\end{equation}
where $t$ indexes time, $\mathcal{A}_h$ is the unknown set of loads in the household, $y_{h,a,t}$ is the power of appliance $a$, and $\varepsilon_{h,t}$ collects measurement and synchronization error. Corpus $d$ provides aggregate windows $\mathbf{x}\in\mathbb{R}^{T_d}$ with $T_d$ samples at cadence $\Delta t_d$, together with submeter trajectories for a subset $\mathcal{L}_h\subset\mathcal{A}_h$. Loads outside $\mathcal{L}_h$ remain in the aggregate as unlabeled mixture components. An appliance observation mask $m^{\mathrm{obs}}_a\in\{0,1\}$ activates its loss only when that submeter exists, and per-sample validity masks remove meter dropouts. REDD and UK-DALE add a unit mismatch because their site meters record \emph{apparent} power and their submeters record \emph{active} power. All training losses operate through these masks.

\paragraph{Prompt-conditioned inverse inference.}
A \emph{request} for category $a$ is a prompt
\begin{equation}
  p_a = \big(\tau_a,\ \mathcal{E}_a\big), \qquad
  \mathcal{E}_a = \{\mathbf{e}_1, \dots, \mathbf{e}_K\},\ K \ge 0,
  \label{eq:prompt}
\end{equation}
where $\tau_a$ is a natural-language appliance card written from generic appliance knowledge, and $\mathcal{E}_a$ is an optional set of \emph{exemplar snippets}: submeter activation traces of category $a$ recorded in other households through public libraries or one-time enrollment. Either component may be absent. A foundation NILM operator is a single parameter set $F_\theta$ that maps an observed window and prompt to a factorized posterior over the requested trajectory:
\begin{equation}
  F_\theta(\mathbf{x}, p_a) = \big(\pi_{a,1:T},\ \mu_{a,1:T},\ \sigma_{a,1:T},\ s_a\big),
  \label{eq:operator}
\end{equation}
Here $\theta$ denotes all trainable parameters, $T$ is the window length, $\pi_{a,t}$ is the activation probability, $\mu_{a,t}$ and $\sigma_{a,t}$ parameterize conditional log-power, and $s_a\in[0,1]$ is the window event score. The default point estimate is
\begin{equation}
  \hat{y}_{a,t}
  =\pi_{a,t}\exp\!\left(\mu_{a,t}+\tfrac{1}{2}\sigma_{a,t}^{2}\right),
  \label{eq:expected-power}
\end{equation}
where $\hat{y}_{a,t}$ is the expected power of requested appliance $a$ at timestep $t$.

\paragraph{Transfer regimes.}
The interface admits three generalization settings. \emph{Held-out household} evaluation draws test windows from periods or households absent from training, other households of the same corpora being supervised. \emph{Category transfer} excludes a category from supervised learning: zero-shot evaluation supplies only its text card ($K{=}0$), few-shot evaluation adds $K$ exemplars from non-test sources, and $\theta$ stays frozen throughout. \emph{Dataset exclusion} removes an entire corpus, with its houses, cadence, voltage regime, and label policy, from supervision, after which enrollment may supply exemplars from its non-test houses at inference.

\paragraph{Evaluation functionals.}
Let $i$ index observed windows for one appliance, let $r_i\in\{0,1\}$ denote center-window activity, and let $s_i$ be the predicted event score. A true activation event is detected when at least one of its overlapping windows has $s_i\geq\eta$, with the default threshold $\eta=0.5$. Positive windows outside true events are merged into temporally contiguous false-alarm episodes. If $N_{\mathrm{TP}}$, $N_{\mathrm{FP}}$, and $N_{\mathrm{FN}}$ denote the resulting numbers of detected events, false-alarm episodes, and missed events, respectively, then
\begin{equation}
  P_{\mathrm{event}}=\frac{N_{\mathrm{TP}}}{N_{\mathrm{TP}}+N_{\mathrm{FP}}},\;
  R_{\mathrm{event}}=\frac{N_{\mathrm{TP}}}{N_{\mathrm{TP}}+N_{\mathrm{FN}}},\;
  F_{1}=\frac{2P_{\mathrm{event}}R_{\mathrm{event}}}
  {P_{\mathrm{event}}+R_{\mathrm{event}}}.
  \label{eq:event-f1}
\end{equation}
$P_{\mathrm{event}}$ and $R_{\mathrm{event}}$ are event precision and recall. For threshold-free window ranking we report the area under the precision--recall curve in average-precision form, $\operatorname{AUPRC}=\sum_{j=1}^{J}(R_j-R_{j-1})P_j$, where $j$ indexes the $J$ distinct score thresholds in descending order and $(P_j,R_j)$ is window-level precision and recall at threshold $j$.

For power metrics, let $c_i$ denote the center sample of window $i$, $\mathcal{I}_{\mathrm{on}}=\{i:r_i=1\}$, and $\mathcal{I}_{\mathrm{off}}=\{i:r_i=0\}$. We report
\begin{equation}
\begin{aligned}
  \operatorname{MAE}_{\mathrm{on}}
    &=\frac{1}{|\mathcal{I}_{\mathrm{on}}|}
      \sum_{i\in\mathcal{I}_{\mathrm{on}}}
      \left|\hat y_{i,c_i}-y_{i,c_i}\right|,\\
  P_{\mathrm{false}}
    &=\frac{1}{|\mathcal{I}_{\mathrm{off}}|}
      \sum_{i\in\mathcal{I}_{\mathrm{off}}}\hat y_{i,c_i},\
  R_{\mathrm{energy}}
    =\frac{\sum_i\sum_{t=1}^{T_i}\hat y_{i,t}}
      {\sum_i\sum_{t=1}^{T_i}y_{i,t}}.
\end{aligned}
\label{eq:power-metrics}
\end{equation}
Here $y_{i,t}$ and $\hat y_{i,t}$ are true and predicted appliance watts, $T_i$ is the number of samples in window $i$, $P_{\mathrm{false}}$ measures hallucinated off-state power, and $R_{\mathrm{energy}}$ is the predicted-to-true energy ratio. Metrics are first averaged over valid appliances within a corpus; the headline result then gives equal weight to the three test corpora.
\section{FM4NILM}
\label{sec:method}

\begin{figure*}[tbp]
\centering
\includegraphics[width=1.0\linewidth]{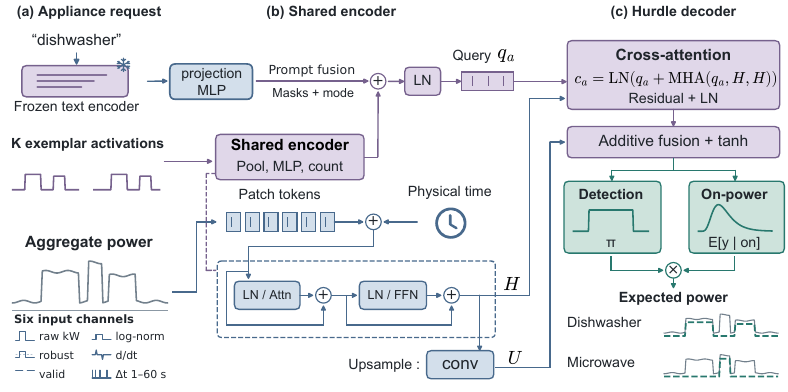}
\caption{\fm: one foundation model for different appliance requests. (a)~Text and optional cross-home exemplars form query $q_a$ through a multilayer perceptron (MLP), modality masking, mode and count embeddings, and layer normalization (LN); the snowflake marks frozen parameters. (b)~Six-channel inputs become patch tokens with physical-time encodings; the inset repeats LN, attention (Attn), and a feed-forward network (FFN) with residual connections six times. The dashed link denotes shared encoder weights; tokens $H$ feed query-to-signal attention, and a convolution (conv) upsamples features $U$. (c)~Activation probability multiplies conditional mean on-power to produce expected power. Gray aggregates and teal predictions show two requests answered by fixed weights. Waveforms are schematic; window scoring and streaming are described in \S\ref{sec:decoder}.}
\Description{Text and exemplar waveforms form a query. Aggregate waveforms become patch tokens, receive physical-time encoding, and pass through six shared attention and feed-forward layers with residual connections. Cross-attention and additive fusion condition two output heads. Activation probability times conditional mean power yields the requested trajectory, illustrated for dishwasher and microwave on the same aggregate.}
\label{fig:arch}
\end{figure*}

Each obstacle named in \S\ref{sec:introduction} becomes one component of
Fig.~\ref{fig:arch}. Cadence heterogeneity is absorbed by a trunk that
tokenizes over physical rather than index time, and that encodes aggregates
and exemplars with the same weights (\S\ref{sec:trunk}). The open appliance
vocabulary is handled by a two-modality encoder that compiles text and
exemplar sets into a single request vector (\S\ref{sec:prompt}), which a
hurdle decoder applies to the aggregate representation, keeping activity and
watt scale on separate heads (\S\ref{sec:decoder}). Partial labeling is
handled in training: a label-free stage first builds the signal
representation (\S\ref{sec:ssl}), after which observation-masked alignment
couples requests to whatever submeters happen to exist (\S\ref{sec:align}).
Heterogeneous pretraining, one shared parameter set, and a compositional
request interface are what make these 3.7M parameters a foundation model
rather than a large multi-task network.

\subsection{Cadence-Aware Signal Trunk}
\label{sec:trunk}

Public corpora sample at 1\,s (REDD, UK-DALE), 8\,s (REFIT, ECO, IDEAL, GREEND), or 60\,s (AMPds2). We retain each native grid and encode physical time explicitly, preserving short events at fine cadence and long context at coarse cadence. The six-channel input at timestep $t$ is
\begin{equation}
\begin{aligned}
  \mathbf{f}_t&=\left[w_t,\ r_t,\ \widetilde w_t,\ d_t,\ v_t,\ \rho_t\right],\\
  r_t&=\frac{\log(1+w_t)}{\log(1+p_{95}^{\mathrm{tr}})},\qquad
  d_t=\frac{w_t-w_{t-1}}{\Delta t_t},\qquad
  \rho_t=\log\frac{\Delta t_t}{\Delta t_d}.
\end{aligned}
\label{eq:input-channels}
\end{equation}
Here $w_t$ is raw power in kilowatts, $r_t$ is log-normalized power, $p_{95}^{\mathrm{tr}}$ is its training-split 95th percentile, $\widetilde w_t$ is robustly standardized power, $d_t$ is the per-second derivative, $v_t\in\{0,1\}$ indicates a valid sample, $\Delta t_t$ is the local sampling interval, and $\Delta t_d$ is the nominal cadence of corpus $d$; all normalization statistics come from the training split.

A width-$P$ strided convolution with patch width $P=4$ embeds $\mathbf f_t$ into tokens, so windows of 480, 120, and 32 samples become 120, 30, and 8 tokens. Each token receives a sinusoidal encoding of its patch center's \emph{physical} offset in seconds, with periods spanning 10\,s--2\,h. A one-minute burst therefore occupies the same temporal frequency band whether sampled at 1\,s or 8\,s. A pre-normalization transformer encoder with six layers, width $D=192$, and four attention heads yields $H\in\mathbb{R}^{L\times D}$, where $L$ is the number of patch tokens. A light convolutional upsampler restores $U\in\mathbb{R}^{T\times D}$, one feature vector for each of the $T$ input samples. The same trunk encodes exemplar snippets, whose mean-pooled tokens form fixed-length embeddings.

\subsection{Two-Modality Prompt Encoder}
\label{sec:prompt}

\paragraph{Text cards.}
Every category has a card set written from generic appliance knowledge: a canonical structured card covering identity, aliases, function, and qualitative electrical behavior, plus paraphrases ranging from a bare name to a colloquial user question. The cards omit dataset, house, and measured statistics. A frozen Qwen3-Embedding model~\cite{zhang2025qwen3embedding} maps them to $\mathbb{R}^{1024}$, and a two-layer multilayer perceptron (MLP) projects the embedding to width $D$. Training samples a random paraphrase at each step, exposing the operator to varied phrasings; evaluation uses the canonical card. The frozen language encoder also accepts requests beyond the training taxonomy.

\paragraph{Exemplar sets.}
An exemplar is a complete activation of the target category inside one submeter window, taken from the training split of a source different from the window under analysis. This cross-domain construction requires the model to use appliance behavior across households. $K$ exemplars are encoded by the trunk, mean-pooled, passed through an MLP, and summed with a count embedding, giving the prompt's grounding vector.

\paragraph{Fusion with modality dropout.}
Text and exemplars form the query
\begin{equation}
  q_a=\operatorname{LN}\!\left(
  m_a^{\mathrm{txt}}W_{\mathrm{txt}}\phi_{\mathrm{txt}}(\tau_a)
  +m_a^{\mathrm{ex}}g(\mathcal{E}_a)
  +e_{\mathrm{mode}}(m_a^{\mathrm{txt}},m_a^{\mathrm{ex}})\right).
  \label{eq:prompt-fusion}
\end{equation}
Here $q_a\in\mathbb{R}^{D}$ is the request vector; $m_a^{\mathrm{txt}},m_a^{\mathrm{ex}}\in\{0,1\}$ indicate whether each modality is present; $\phi_{\mathrm{txt}}(\tau_a)\in\mathbb{R}^{1024}$ is the frozen language encoding of card $\tau_a$; and $W_{\mathrm{txt}}$ projects that encoding to width $D$. The function $g(\mathcal{E}_a)$ is the MLP projection of the encoded exemplar set, $e_{\mathrm{mode}}$ is a learned embedding of the modality combination, and $\operatorname{LN}$ denotes layer normalization. During training, a request uses both modalities with probability 40\%, text alone with probability 40\%, and exemplars alone with probability 20\%. This schedule calibrates one operator for every supported prompt configuration and makes exemplars an \emph{in-context learning} channel.

\subsection{Request-Conditioned Hurdle Decoder}
\label{sec:decoder}

The query retrieves relevant trunk tokens through multi-head attention (MHA) and then conditions every timestep:
\begin{equation}
\begin{aligned}
  c_a&=\operatorname{LN}\!\left(q_a+
    \operatorname{MHA}(q_a,H,H)\right),\\
  r_{a,t}&=\tanh\!\left(W_hU_t+W_qc_a\right).
\end{aligned}
\label{eq:decoder-fusion}
\end{equation}
Here $c_a\in\mathbb{R}^{D}$ is the request-conditioned context, $H$ supplies attention keys and values, $U_t$ is the upsampled signal feature at timestep $t$, $W_h$ and $W_q$ are learned projections, and $r_{a,t}$ is the fused feature. All decoder weights are shared across requests; $q_a$ is the target selector.

Three learned linear heads parameterize the hurdle distribution from $r_{a,t}$: the activation logit $\ell_{a,t}=w_{\pi}^{\top}r_{a,t}+\beta_{\pi}$ gives activity probability $\pi_{a,t}=\sigmoid(\ell_{a,t})$; base power $b_{a,t}=\softplus(w_b^{\top}r_{a,t}+\beta_b)$ sets the conditional log-power mean $\mu_{a,t}=\log(b_{a,t}+10^{-6})$; and $\softplus(w_{\sigma}^{\top}r_{a,t}+\beta_{\sigma})$, clipped to $[0.05,1.5]$, gives the scale $\sigma_{a,t}$. These define the Bernoulli--lognormal hurdle:
\begin{equation}
\begin{aligned}
  z_{a,t}&\sim\operatorname{Bernoulli}(\pi_{a,t}),\
  y_{a,t}\mid z_{a,t}{=}0=0,\\
  \log y_{a,t}\mid z_{a,t}{=}1
    &\sim\mathcal{N}(\mu_{a,t},\sigma_{a,t}^{2}),\
  \hat y_{a,t}=\pi_{a,t}\exp\!\left(
    \mu_{a,t}+\tfrac{1}{2}\sigma_{a,t}^{2}\right).
\end{aligned}
  \label{eq:hurdle}
\end{equation}
Here $z_{a,t}$ is the binary activity state, $y_{a,t}$ is conditional appliance power, $\mathcal{N}$ is a normal distribution in log-power space, and $\hat y_{a,t}$ is expected power. The hurdle factorization assigns sparse-event ranking and positive-watt calibration to separate heads and losses.

The window score combines timestep evidence with request context. Normalized log-sum-exp evidence across $T$ timesteps, $\bar\ell_a=\tau[\log\sum_{t=1}^{T}\exp(\ell_{a,t}/\tau)-\log T]$ with temperature $\tau$ clipped to $[0.5,8]$, feeds a learned pooling head $s_a=\sigmoid(\alpha\bar\ell_a+w_s^{\top}c_a+\beta_s)$, tying window ranking to local activity logits while retaining a context correction from $c_a$.

\paragraph{Streaming inference.}
A sliding meter produces overlapping windows. For window $i$ starting at $u_i$, let $\mathcal{N}_i=\{j:-P\le u_j-u_i\le F\}$ collect same-household windows within past support $P$ and future support $F$. Streaming replaces the event score with a triangular-weighted average $\widetilde s_{a,i}=\sum_{j\in\mathcal{N}_i}\omega_{ij}s_{a,j}/\sum_{j\in\mathcal{N}_i}\omega_{ij}$, $\omega_{ij}=\max(0,1-|u_j-u_i|/(\max(P,F)+\Delta t_d))$. Zero-lookahead uses $(P,F)=(120\,\mathrm{s},0)$ and bounded delay $(120\,\mathrm{s},60\,\mathrm{s})$; both reuse frozen scores and add no parameters. Per-timestep power is left unpooled, since averaging trajectories suppresses active peaks (\S\ref{sec:results-deployment}).

\subsection{Masked Pretraining over Heterogeneous Corpora}
\label{sec:ssl}

Every train-split window contributes its aggregate and observed submeter channels as independent univariate sequences, producing 645k sequences across seven corpora. We replace 40\% of patch tokens with a mask token and optimize
\begin{equation}
  \mathcal{L}_{\mathrm{pre}}=
  \frac{\sum_{q=1}^{L}M_qV_q\,
    \operatorname{Huber}(\widehat r_q-r_q)}
       {\sum_{q=1}^{L}M_qV_q}.
  \label{eq:pretraining}
\end{equation}
Here $q$ indexes patch tokens, $M_q\in\{0,1\}$ marks a masked token, $V_q\in\{0,1\}$ a valid one, $r_q$ is the log-normalized target, and $\widehat r_q$ the trunk reconstruction; Huber is quadratic near zero and linear on large residuals. The stage needs no category labels. Aggregate channels expose mixtures and submeter channels expose the isolated signatures later supplied as exemplars, so both halves of the request interface are represented before any supervision. Held-out variants remove the relevant category or corpus channels from this pool.

\subsection{Episodic Partial-Label Alignment}
\label{sec:align}

Supervised training interleaves single-source batches.  Each step samples an event-balanced batch (40\% rare-category event windows, 20\% common-category positives, 20\% high-activity hard negatives, 20\% natural), builds one prompt per requested category under modality dropout, and optimizes
\begin{equation}
\begin{aligned}
  \mathcal{L} = {} & 5\,\mathcal{L}_{\mathrm{on}} + 0.2\,\mathcal{L}_{\mathrm{off}} + 0.3\,\mathcal{L}_{\mathrm{state}} + 0.2\,\mathcal{L}_{\mathrm{energy}} \\
  & + 0.2\,\mathcal{L}_{\mathrm{peak}} + 0.1\,\mathcal{L}_{\mathrm{NLL}} + 10^{-4}\,\mathcal{L}_{\mathrm{TV}},
\end{aligned}
\label{eq:loss}
\end{equation}
Each term has a distinct role: $\mathcal{L}_{\mathrm{on}}$ is SmoothL1 error on conditional watts at active timesteps, $\mathcal{L}_{\mathrm{off}}$ penalizes squared expected power at inactive ones, and $\mathcal{L}_{\mathrm{state}}$ is positive-reweighted binary cross-entropy (BCE) at timestep and window levels; $\mathcal{L}_{\mathrm{energy}}$ and $\mathcal{L}_{\mathrm{peak}}$ preserve event energy and peak scale, $\mathcal{L}_{\mathrm{NLL}}$ is the conditional-lognormal negative log-likelihood (NLL), and $\mathcal{L}_{\mathrm{TV}}$ is total variation (TV) regularization against spurious switching. Crucially, observation and validity masks multiply every term, so an unmetered load contributes exactly zero gradient rather than a spurious zero target. This is what makes partially labeled corpora trainable at all. Coefficients are fixed on development data, and checkpoints are selected by mean validation AUPRC over included sources under text-only canonical prompts.

\paragraph{Held-out protocols.}
Category transfer excludes the held-out category from supervised requests, event-balanced sampling, checkpoint selection, and the pretraining submeter pool. At test time, the model receives its text card and optional exemplars. Dataset transfer excludes the target corpus from supervised training; strict exclusion also removes it from pretraining. A second variant includes its unlabeled aggregate signals during pretraining.
\section{Experimental Setup}
\label{sec:setup}

\paragraph{Corpora and frozen splits.}
Seven public corpora participate (Table~\ref{tab:corpus-stats}): REDD~\cite{kolter2011redd}, UK-DALE~\cite{kelly2015ukdale}, REFIT~\cite{murray2017refit}, ECO~\cite{beckel2014eco}, IDEAL~\cite{pullinger2021ideal}, GREEND~\cite{monacchi2014greend}, and AMPds2~\cite{makonin2016ampds2}. All seven supply supervision, masked pretraining data, and exemplars to the main model at their native cadences; REDD, UK-DALE, and REFIT additionally carry the frozen test splits, and the three-corpus ablation is supervised by those three alone.

One split is used throughout. REDD houses 1--2 and UK-DALE houses 1--2 follow chronological 70/10/20 partitions separated by purge gaps, with realized boundary gaps of 22--34 hours; REFIT is partitioned by house (3/5/6/9 train, 2/11 validation, 13/20 test). Purge gaps are applied before overlapping windows are generated, and direct sample-index checks confirm no shared samples between train and test covers. The test set holds 18{,}912 windows: 7{,}332 REDD, 6{,}580 UK-DALE, 5{,}000 REFIT. Checkpoint selection and threshold calibration use validation data only.

\paragraph{Aggregate provenance.}
Six corpora use the site-meter channels identified in their metadata. GREEND lacks a whole-home channel, so its aggregate sums all available plug meters; GREEND contributes training data and has no test split. Table~\ref{tab:provenance} quantifies each mixture: observed submeters explain only 6--35\% of aggregate power, and REFIT's 6.2\% labeled share leaves an average 459\,W of unobserved load. Every requested appliance must therefore be separated from substantial unmeasured demand.

\paragraph{Metrics and baselines.}
One implementation scores every model on window event F1 at a fixed 0.5 operating point, AUPRC, MAE-on, false power, and energy ratio (\S\ref{sec:problem}). We train two baseline families on the same windows. The specialist family assigns one model to each appliance--corpus pair, for twelve deployed models. It includes factorial HMM~\cite{kolter2012fhmm}, gradient-boosted trees~\cite{chen2016xgboost}, convolutional Seq2Point and Seq2Seq~\cite{zhang2018sequence}, recurrent BiLSTM~\cite{kelly2015neural}, and transformer BERT4NILM~\cite{yue2020bert4nilm} and NILMFormer~\cite{petralia2025nilmformer}. All receive the same event-balanced training budget, with predicted center power as their on/off score. The shared-network family uses fixed per-appliance output heads at matched recurrent width and parameter count, isolating the effect of replacing a closed output inventory with a request interface.

\paragraph{Training and matrix.}
The trunk is pretrained for 20 epochs over 645k sequences with span masking on the raw-kilowatt target (about eight minutes on one graphics processing unit (GPU)). Supervised alignment runs for 60 epochs of 400 event-balanced steps with batch size 128, AdamW, a one-cycle learning rate of $3{\times}10^{-4}$, and bfloat16 (bf16) automatic mixed precision (about 25 minutes on one GPU). Every configuration uses five seeds; we report mean $\pm$ standard deviation and paired per-seed differences.

The study spans the full seven-corpus model; ablations of pretraining, text, exemplar episodes, amplitude evidence, and supervision breadth; category transfer excluding microwave or dishwasher everywhere, including from the pretraining submeter pool; dataset transfer excluding REFIT under strict and unlabeled-target settings; and input interventions that swap text cards or exemplars for mismatched alternatives or zero the aggregate. In total we train over 80 models and evaluate more than 200 held-out configurations on eight NVIDIA RTX 5090 GPUs.

\begin{table}[!tbp]
\centering
\caption{Aggregate provenance and label coverage. In the header, Lab., Agg., Res., and Exc. denote labeled share, aggregate power, residual power, and exceeds, respectively. Labeled share is the mean ratio of observed submeter power to aggregate power. Residual is the remaining consumption, and exceeds is the fraction of samples where the labeled sum is larger than the independently measured aggregate.}
\label{tab:provenance}
\small
\setlength{\tabcolsep}{2pt}
\begin{tabular*}{\linewidth}{@{\extracolsep{\fill}}llrrrr@{}}
\toprule
Corpus & Aggregate & Lab.\ (\%) & Agg.\ (W) & Res.\ (W) & Exc.\ (\%) \\
\midrule
REDD & site meter & 30.6 & 309 & 205 & 0.08 \\
UK-DALE & site meter & 19.7 & 362 & 243 & 0.17 \\
REFIT & site meter & 6.2 & 543 & 459 & 0.74 \\
ECO & site meter & 30.1 & 417 & 305 & 0.47 \\
IDEAL & site meter & 25.3 & 485 & 346 & 2.23 \\
GREEND & synth.\ (all meters) & 34.6 & 257 & 142 & 0.00 \\
AMPds2 & site meter & 7.1 & 1271 & 1153 & 0.00 \\
\bottomrule
\end{tabular*}
\end{table}
\section{Results}
\label{sec:results}

We pose progressively harder requests. Can one \fm model serve
\emph{familiar} requests as well as a fleet of specialists, and which
ingredients earn that result (\S\ref{sec:results-main}--\ref{sec:results-scaling})?
Does it answer requests for appliances and corpora it was never supervised
on, and is it truly reading the request when it does
(\S\ref{sec:results-zeroshot}--\ref{sec:results-controls})? Finally, where
does the interface run out, and what does deployment context buy back
(\S\ref{sec:results-containment}--\ref{sec:results-deployment})? Unless
stated otherwise, comparisons use five seeds, each seed weights the three
test corpora equally, and intervals are bootstrapped over paired per-seed
differences.

\subsection{Accuracy on Held-Out Households}
\label{sec:results-main}

Generality is usually bought with accuracy; Table~\ref{tab:main} shows that
here it is not. One text-prompted \fm model serves all twelve
appliance--corpus requests at 55.6\% event F1, 62.5\% AUPRC, and 251.8\,W
MAE-on. BERT4NILM, the strongest specialist detector, reaches 57.1\% F1 and
63.0\% AUPRC with a dedicated network per row: twelve models buy 1.5 points
of F1 over one, while \fm cuts active-window watt error by 72.6\,W against that
baseline and 169.1\,W against XGBoost. The shared model trails on detection
by roughly seed noise and leads on power by a margin that is not.

The learned-identity (ID) control isolates what language contributes.
Replacing cards with a free appliance embedding, everything else fixed,
drops AUPRC to 49.4\% and, more diagnostically, inflates its seed standard
deviation from 2.2 to 18.6 points. An index must discover the relationships
among appliances from disaggregation supervision alone; a frozen language
encoder supplies them, so electrically similar requests start near one
another and training becomes reproducible rather than lucky. Adding five
exemplars leaves familiar accuracy intact while arming the same interface
for the transfer settings below.

\begin{table*}[!tbp]
\centering
\caption{Held-out test results over 18{,}912 windows from unseen houses and periods using five seeds. Specialist baselines use one model per appliance and corpus, totaling twelve models. The closed-vocabulary control replaces the text prompt with a learned appliance identity. \fm uses one shared, text-addressed network. ``Deployed'' gives the total parameter count for all twelve appliance--corpus requests. Best values are bold.}
\label{tab:main}
\small
\setlength{\tabcolsep}{6pt}
\begin{tabular*}{\linewidth}{@{\extracolsep{\fill}}lrrrr@{}}
\toprule
Model & Deployed & Event F1 (\%) & AUPRC (\%) & MAE-on (W) \\
\midrule
\multicolumn{5}{l}{\emph{Appliance-specific, source-local (12 models)}} \\
FHMM~\cite{kolter2012fhmm} & 12 $\cdot$ tiny & 34.9 $\pm$ 5.9 & 18.5 $\pm$ 2.5 & 661.3 $\pm$ 53.7 \\
XGBoost~\cite{chen2016xgboost} & 12 $\cdot$ trees & 51.6 $\pm$ 0.6 & 59.3 $\pm$ 1.1 & 420.9 $\pm$ 13.3 \\
Seq2Point~\cite{zhang2018sequence} & 222\,M & 47.3 $\pm$ 1.3 & 44.4 $\pm$ 0.7 & 878.0 $\pm$ 3.5 \\
Seq2Seq~\cite{zhang2018sequence} & 226\,M & 53.5 $\pm$ 0.8 & 59.5 $\pm$ 1.5 & 563.8 $\pm$ 15.1 \\
BiLSTM~\cite{kelly2015neural} & 6.9\,M & 54.9 $\pm$ 1.6 & 61.5 $\pm$ 2.5 & 325.2 $\pm$ 18.8 \\
BERT4NILM~\cite{yue2020bert4nilm} & 7.3\,M & \textbf{57.1 $\pm$ 2.0} & 63.0 $\pm$ 2.0 & 324.4 $\pm$ 24.4 \\
NILMFormer~\cite{petralia2025nilmformer} & 7.2\,M & 46.4 $\pm$ 1.6 & 51.3 $\pm$ 2.0 & 525.0 $\pm$ 18.0 \\
\midrule
\multicolumn{5}{l}{\emph{Shared network, one model (ours)}} \\
Closed vocab.\ (learned ID) & 3.7\,M & 49.6 $\pm$ 5.3 & 49.4 $\pm$ 18.6 & 431.0 $\pm$ 172.8 \\
\fm (text prompt) & 3.7\,M & 55.6 $\pm$ 0.4 & 62.5 $\pm$ 2.2 & \textbf{251.8 $\pm$ 6.3} \\
\fm (text + 5 ex.) & 3.7\,M & 56.2 $\pm$ 0.9 & 62.4 $\pm$ 2.2 & 259.2 $\pm$ 9.3 \\
\bottomrule
\end{tabular*}
\end{table*}

\paragraph{Validation-calibrated operating point.}
The fixed 0.5 threshold assumes no target-domain labels, the realistic
deployment posture. One validation-selected threshold per appliance raises
\fm event F1 to 57.0\% ($+1.4$ points [+0.8, +2.2]), level with BERT4NILM, so
the residual gap at that threshold is an operating-point artifact rather than
a ranking deficit. Its distribution is informative: $+11.3$ points on
UK-DALE, $-0.7$ on REDD, $-6.4$ on REFIT. REDD and UK-DALE hold out periods from validation
\emph{households}; REFIT holds out entire households. A calibrated threshold
travels across time within a home but not across homes. It is the first
household-specific quantity the request cannot carry, and a preview of
\S\ref{sec:results-scale}.

\subsection{Attribution: What Each Ingredient Buys}
\label{sec:results-ablation}

Table~\ref{tab:ablation} traces that accuracy back to the recipe, whose
components split into those that buy detection and those that buy watts.
Removing span-masked pretraining costs 1.3 points of event F1, 2.2 of AUPRC,
and 17\,W with paired intervals excluding zero: the label-free stage earns its
place though it never sees an appliance name. Removing exemplar episodes
costs 2.7 points of AUPRC on an interval spanning zero but a decisive 118\,W of
MAE-on, since training with exemplars grounds \emph{amplitude} even when the
deployed request is text-only. It also forfeits the few-shot mechanism of
\S\ref{sec:results-zeroshot} outright. Broadening supervision from three to
seven corpora adds 2.5 points of event F1 and removes 35\,W at similar AUPRC, so
corpora the model is never tested on still improve the ones it is.
Role-separated amplitude features trade equal AUPRC for higher watt error
and were not adopted; the learned-ID arm is the table's largest single
regression.

\begin{table}[!tbp]
\centering
\caption{Five-seed ablations (text-only prompts). $\Delta$AUPRC is the paired per-seed mean against the full \fm model with a bootstrap 95\% confidence interval (CI); an interval excluding zero is significant.}
\label{tab:ablation}
\small
\setlength{\tabcolsep}{0pt}
\begin{tabular*}{\linewidth}{@{\extracolsep{\fill}}lrrr@{}}
\toprule
Variant & Event F1 (\%) & AUPRC (\%) & MAE-on (W) \\
\midrule
\fm (full) & 55.6 $\pm$ 0.4 & 62.5 $\pm$ 2.2 & 251.8 $\pm$ 6.3 \\
$-$ masked pretraining & 54.3 $\pm$ 0.9 & 60.3 $\pm$ 1.8 & 269.1 $\pm$ 8.5 \\
$-$ text (learned ID) & 49.6 $\pm$ 5.3 & 49.4 $\pm$ 18.6 & 431.0 $\pm$ 172.8 \\
$-$ exemplar episodes & 53.3 $\pm$ 1.2 & 59.8 $\pm$ 2.6 & 369.6 $\pm$ 24.4 \\
$+$ amplitude roles & 54.1 $\pm$ 1.6 & 63.6 $\pm$ 1.1 & 297.8 $\pm$ 16.3 \\
three corpora & 53.1 $\pm$ 1.3 & 63.5 $\pm$ 1.1 & 287.2 $\pm$ 11.7 \\
\midrule
\multicolumn{4}{l}{\emph{Paired $\Delta$AUPRC in points vs.\ full model [95\% CI]:}} \\
\multicolumn{4}{l}{$-$pretraining $-2.2$ $[-3.9, -0.5]$} \\
\multicolumn{4}{l}{$-$exemplar episodes $-2.7$ $[-5.5, +0.7]$} \\
\multicolumn{4}{l}{three corpora $+1.0$ $[-1.8, +3.9]$} \\
\multicolumn{4}{l}{amplitude roles $+1.1$ $[-0.7, +3.1]$} \\
\bottomrule
\end{tabular*}
\end{table}

\subsection{Model and Pretraining Scale}
\label{sec:results-scaling}

\begin{figure*}[tbp]
\centering
\includegraphics[width=0.84\linewidth]{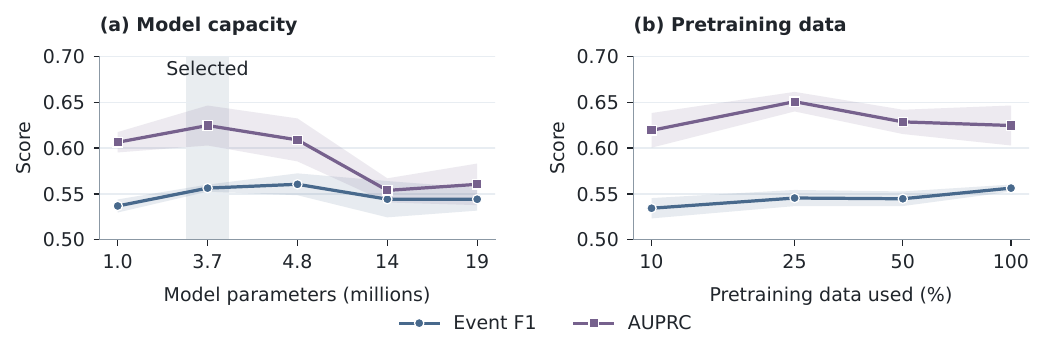}
\caption{Model and pretraining scale. Points show five-seed means; bands show $\pm 1$ standard deviation. \textbf{(a)}~Detection scores across model sizes, spaced categorically; the shaded column marks the selected 3.7M model. \textbf{(b)}~Event F1 improves overall as the pretraining pool grows from 10\% to 100\%, while AUPRC peaks at 25\%.}
\Description{Two line charts show event F1 and AUPRC against model parameter count and pretraining-data fraction. The selected model has 3.7 million parameters.}
\label{fig:scaling}
\end{figure*}

If pretraining helps, which axis of scale should one spend on? Fig.~\ref{fig:scaling} answers unambiguously. Width and depth sweeps under matched schedules place the selected 3.7M-parameter trunk at the accuracy frontier; larger trunks degrade watt accuracy without a consistent detection gain, since the supervision available from partially labeled homes does not grow with them. Pretraining data behaves differently: enlarging the unlabeled pool from 10\% to 100\% improves event F1 overall, with AUPRC peaking at 25\%. NILM is not yet capacity-limited but corpus-limited. That is an encouraging diagnosis, since unlabeled aggregate streams are free to a metering fleet and submetered homes are not.

\subsection{Zero-Shot and Few-Shot Category Transfer}
\label{sec:results-zeroshot}

\begin{figure*}[tbp]
\centering
\includegraphics[width=0.74\linewidth]{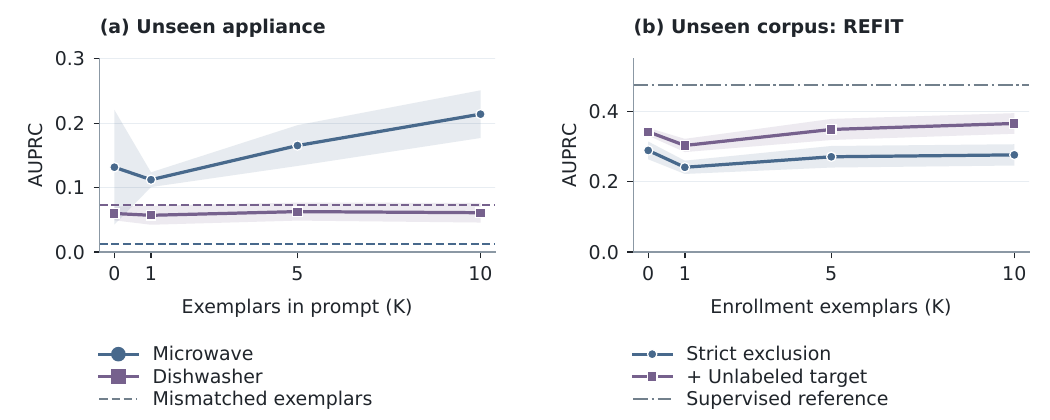}
\caption{Transfer through exemplar requests. Curves show five-seed mean AUPRC with $\pm 1$ standard deviation; $K{=}0$ denotes text alone. \textbf{(a)}~Microwave improves from one to ten exemplars, while dishwasher stays near its mismatched-exemplar control. Dashed references use five mismatched exemplars with text omitted. \textbf{(b)}~REFIT transfer under strict exclusion and with unlabeled target pretraining. The dash-dot line is the supervised REFIT reference.}
\Description{AUPRC versus exemplar count for excluded microwave and dishwasher categories and for the excluded REFIT corpus. Each appliance has its own mismatched-exemplar reference. REFIT curves compare strict exclusion with unlabeled target pretraining.}
\label{fig:incontext}
\end{figure*}

We now ask for an appliance that no part of training ever described. The
held-out category is purged from supervised requests, event-balanced
sampling, checkpoint selection, and the pretraining submeter pool; the
frozen network then sees only its text card and, optionally, exemplars from
other households. For microwave the card alone reaches 13.2\%
AUPRC, above chance but only just (Table~\ref{tab:loao}). Ten exemplars
carry it to 21.4\%, a paired gain of $+10.2$ points over one exemplar, and
remove 189.5\,W of MAE-on (Fig.~\ref{fig:incontext}a). Nothing in the network changed: a category the weights
never encoded became measurable purely through the request.

\begin{figure}[tbp]
\centering
\includegraphics[width=0.9\linewidth]{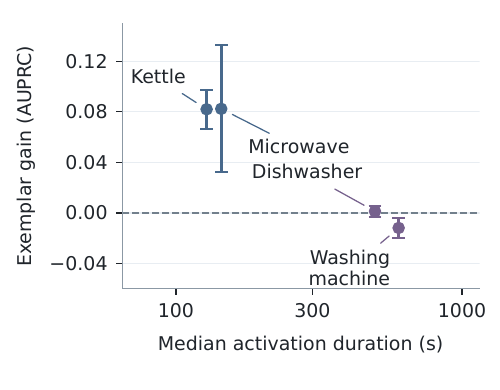}
\caption{Exemplar benefit and activation duration. Each point is the paired AUPRC gain from ten exemplars over text alone, averaged across five seeds; bars show $\pm 1$ standard error. Duration is the median from submeter activations, shown on a log scale. Kettle and microwave show the largest gains.}
\Description{A scatter plot relates exemplar AUPRC gain to median appliance activation duration; short microwave and kettle events show larger gains than long multi-phase loads.}
\label{fig:spectrum}
\end{figure}

The dishwasher refuses to follow, and the refusal is the more informative
result. It stays near 6\% AUPRC in every prompt mode, and a
synthetic-mixture probe supplying abundant, exactly labeled dishwasher
activations during training still yields only $+1.8$ points from one to ten
exemplars, so the obstacle is not exemplar scarcity.
Fig.~\ref{fig:spectrum} identifies what it is: ordered by median activation
duration, ten exemplars add 8.2 points of AUPRC for kettle and microwave,
$-1.2$ for washing machine, and essentially nothing for dishwasher. An
exemplar is a fixed-length window and can only transmit a signature that
\emph{fits inside one}. A microwave burst, median 36--80\,s across our
corpora, is contained whole; REFIT's median dishwasher cycle of 4112\,s is
glimpsed only in fragments that are individually ambiguous with other loads
(Appendix Table~\ref{tab:window-span}). Few-shot capacity is bounded not by
the number of examples but by the ratio of activation duration to prompt
span, a prediction that \S\ref{sec:results-deployment} tests directly by
widening the span instead of adding exemplars.

\begin{table}[!tbp]
\centering
\caption{Zero-/few-shot transfer to categories excluded from all supervision and from pretraining submeter channels.  Dense held-out test windows; five seeds $\times$ five exemplar draws.  The held-out category is absent from supervision, checkpoint selection, and the pretraining submeter pool.}
\label{tab:loao}
\small
\setlength{\tabcolsep}{2pt}
\begin{tabular*}{\linewidth}{@{\extracolsep{\fill}}llrrr@{}}
\toprule
Category & Prompt & Event F1 (\%) & AUPRC (\%) & MAE-on (W) \\
\midrule
\multirow{5}{*}{Microwave}
 & text only & 31.1 $\pm$ 9.3 & 13.2 $\pm$ 8.9 & 968.4 $\pm$ 140.5 \\
 & text + 1 ex. & 24.5 $\pm$ 2.2 & 11.2 $\pm$ 1.1 & 1113.6 $\pm$ 35.6 \\
 & text + 5 ex. & 28.1 $\pm$ 4.1 & 16.5 $\pm$ 3.2 & 1021.8 $\pm$ 45.2 \\
 & text + 10 ex. & 36.5 $\pm$ 4.7 & 21.4 $\pm$ 3.7 & 924.1 $\pm$ 52.1 \\
 & 5 ex.\ only & 27.5 $\pm$ 2.8 & 17.1 $\pm$ 2.3 & 1026.0 $\pm$ 33.2 \\
\midrule
\multirow{5}{*}{Dishwasher}
 & text only & 12.8 $\pm$ 0.7 & 6.0 $\pm$ 1.1 & 1013.3 $\pm$ 48.3 \\
 & text + 1 ex. & 12.0 $\pm$ 1.1 & 5.7 $\pm$ 1.5 & 1061.7 $\pm$ 41.6 \\
 & text + 5 ex. & 12.2 $\pm$ 2.4 & 6.3 $\pm$ 1.4 & 1025.7 $\pm$ 50.9 \\
 & text + 10 ex. & 12.0 $\pm$ 2.2 & 6.1 $\pm$ 1.5 & 1033.1 $\pm$ 51.0 \\
 & 5 ex.\ only & 11.8 $\pm$ 1.3 & 6.4 $\pm$ 1.2 & 1022.0 $\pm$ 44.5 \\
\bottomrule
\end{tabular*}
\end{table}

\subsection{Dataset Exclusion and Enrollment}
\label{sec:results-lodo}

Category transfer changes the requested component but keeps the meter domain
familiar. We now remove an entire domain: REFIT, with its houses, cadence,
voltage regime, and labeling policy. What must an operator supply to
recover it? The answer forms a ladder ordered by acquisition cost.
Supervision from the auxiliary 8\,s and 60\,s corpora alone, requiring
nothing from REFIT, already reaches 28.9\% AUPRC, 61\% of the 47.5\%
supervised reference (Table~\ref{tab:lodo}). Admitting REFIT's
\emph{unlabeled} aggregates to pretraining, a stream any deployed meter
emits for free, lifts text-only AUPRC to 34.2\%. Only then does labeled
enrollment contribute, modestly: five target-domain examples reach 34.9\%,
against 30.1\% for cross-source examples under strict exclusion
(Fig.~\ref{fig:incontext}b). What a new
corpus mostly costs is not its labels but its cadence.

Holding out UK-DALE reproduces the pattern: text reaches 33.1\% AUPRC, 51\%
of its supervised reference, and ten enrollment examples raise it to 35.8\%.
Two auxiliary levers behave less simply. Synthetic-mixture augmentation
raises REFIT text-only AUPRC from 34.2\% to 39.4\% while moving in-domain
AUPRC by $-2.4$ points [$-4.3$, $-0.5$], a tradeoff a deployment can tune
but not escape. And text-guided exemplar retrieval matches random enrollment for
supervised categories yet loses 8.5 points of AUPRC for the unseen microwave,
exactly where curation would matter most: choosing exemplars for an unseen
category is itself an open problem.

\begin{table}[!tbp]
\centering
\caption{Dataset exclusion (REFIT test houses 13/20). Strict exclusion removes REFIT from supervision and pretraining; the unlabeled-target variant admits only its unlabeled aggregates to pretraining. Enrollment draws exemplars from REFIT non-test houses at inference. MAE-on is in the released per-run JavaScript Object Notation (JSON) files.}
\label{tab:lodo}
\small
\setlength{\tabcolsep}{4pt}
\begin{tabular*}{\linewidth}{@{\extracolsep{\fill}}llrr@{}}
\toprule
Training & Prompt & Event F1 (\%) & AUPRC (\%) \\
\midrule
REFIT trained (reference) & text & 32.2 $\pm$ 2.2 & 47.5 $\pm$ 6.3 \\
\midrule
\multirow{4}{*}{Strict}
 & text & 23.4 $\pm$ 2.0 & 28.9 $\pm$ 2.5 \\
 & + 1 enroll. & 20.6 $\pm$ 2.6 & 24.1 $\pm$ 1.9 \\
 & + 5 enroll. & 21.2 $\pm$ 3.3 & 27.1 $\pm$ 3.1 \\
 & + 5 cross-src. & 23.8 $\pm$ 2.5 & 30.1 $\pm$ 2.3 \\
\midrule
\multirow{2}{*}{+ unlab.\ target}
 & text & 29.4 $\pm$ 2.1 & 34.2 $\pm$ 1.1 \\
 & + 5 enroll. & 26.8 $\pm$ 1.6 & 34.9 $\pm$ 3.0 \\
\bottomrule
\end{tabular*}
\end{table}

\subsection{Causal Controls on the Prompt}
\label{sec:results-controls}

\begin{figure}[tbp]
\centering
\includegraphics[width=0.9\linewidth]{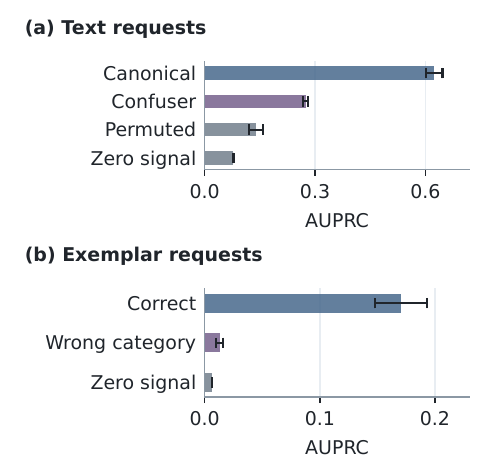}
\caption{Input interventions. Bars show five-seed mean AUPRC with $\pm 1$ standard deviation. \textbf{(a)}~Text requests across all evaluated appliance--corpus pairs. \textbf{(b)}~Five exemplars without text for the excluded microwave. Altering the request or removing the aggregate signal reduces accuracy; panels use different AUPRC scales.}
\Description{Bar charts compare canonical prompts with altered text, missing aggregate input, and wrong-category exemplars; each intervention reduces AUPRC.}
\label{fig:controls}
\end{figure}

A prompt-conditioned model invites an obvious suspicion: that the prompt is
merely a lookup key and the network a bank of heads in disguise.
Fig.~\ref{fig:controls} intervenes on each input in turn, and the
\emph{graded} damage rules this out. An electrically matched confuser card
lowers AUPRC from 62.5\% to 27.4\%, whereas a fixed derangement of the
card--appliance assignment lowers it further to 13.8\%. A lookup key would
fail identically under both; a model reading semantic content degrades more
gently when the wrong card still describes a similar load, as observed. Keeping the correct card
but zeroing the aggregate collapses accuracy to 7.8\%, so the request
selects a component of the measurement rather than synthesizing one from
priors. The exemplar channel behaves alike: for the held-out microwave,
substituting another category's exemplars drops exemplar-only AUPRC from
17.1\% to 1.3\%. Language, exemplars, and the aggregate therefore each
exert independent, necessary control over the output; Appendix
Table~\ref{tab:controls} lists the exact values.

\subsection{Scale Containment}
\label{sec:results-containment}

Selecting the right component is necessary but not sufficient: applications
integrate the trajectory rather than thresholding it, so the estimate must
stay contained while the appliance is off. MAE-on
scores only active windows, so Table~\ref{tab:containment} adds off-state
false power and the predicted-to-true energy ratio. On held-out households
\fm emits 31.5\,W of false power at a 2.54 energy ratio, against 37.5\,W and
4.35 for the learned-ID control, and 20.7\,W and 1.29 under strict dataset
exclusion. This is the Bernoulli gate doing work a single regression head
cannot: it drives expected power to zero multiplicatively, whereas a
regressor must learn to emit a small number and inevitably leaks. It cannot
fix the error surviving \emph{while} the appliance runs.

\begin{table}[!tbp]
\centering
\small
\caption{Scale containment. False power is the mean predicted wattage while the requested appliance is off; energy ratio is predicted energy divided by true energy. Their ideal values are zero and one, respectively.}
\label{tab:containment}
\setlength{\tabcolsep}{2pt}
\begin{tabular*}{\linewidth}{@{\extracolsep{\fill}}lrr@{}}
\toprule
Setting & False power (W) & Energy ratio \\
\midrule
\fm (held-out households) & 31.5 $\pm$ 2.5 & 2.54 $\pm$ 0.08 \\
Closed vocab. (ID) & 37.5 $\pm$ 22.8 & 4.35 $\pm$ 3.08 \\
\fm, unseen (strict) & 20.7 $\pm$ 4.1 & 1.29 $\pm$ 0.23 \\
\fm, unseen (+unlab.) & 24.1 $\pm$ 5.8 & 1.54 $\pm$ 0.35 \\
\bottomrule
\end{tabular*}
\end{table}

\subsection{Power Error Across Households}
\label{sec:results-scale}

To ask \emph{what kind} of error remains, we fit one diagnostic multiplier
per household--appliance group from test labels (an oracle, used only for
attribution) and measure how much MAE-on it removes: what the multiplier
absorbs is level error, what survives is trajectory shape. (This diagnostic
weights 14 groups equally rather than three corpora, hence 202\,W where
Table~\ref{tab:main} reads 251.8\,W.) Table~\ref{tab:scale-decomp} attributes
19\% of group-macro MAE to absolute scale under full supervision and 25\%
once REFIT is held out, where the median multiplier rises from 1.25 to 2.11
and \emph{every} held-out group is under-predicted, a systematic bias rather
than dispersion.

What a cross-corpus request carries explains that bias. Median on-power
across test households spans 7.1$\times$ for dishwashers and 2.3$\times$ for
refrigerators, so identical trajectory shapes sit at very different watt
levels depending on nameplate rating, operating mode, and supply voltage. A
card describes a category and a foreign-household exemplar describes that
category's shape; neither observes the rating of the specific unit behind
this meter, so the model regresses toward the training population's level.
\emph{The request transfers an appliance's pattern, not its level}. That is
the sharpest characterization of this interface's current reach.

\paragraph{Explicit scale conditioning.}
We tried four ways of writing the missing level into the request: bounded
scale terms, power priors in the card, exemplar-amplitude summaries, and
aggregate-relative watt heads. All fail for one reason. A request-level
scale term \emph{increases} in-domain MAE-on by 6 [2, 11]\,W; exemplar
amplitude moves held-out AUPRC from 36.6\% to 34.7\% with no consistent watt
gain, because a cross-corpus exemplar's amplitude is drawn from another
household and is nearly uninformative about this one. The single
intervention that helps points the same way: sourcing half the training
exemplars from the same corpus lowers held-out MAE-on from 761 to 678\,W at
ten exemplars in four of five seeds, though the paired interval includes
zero and in-domain F1 moves by $-1.1$ points [$-2.6$, $-0.0$]. Level, unlike
shape, is not transferable information; it must be measured where it is used.

\begin{table}[tbp]
\centering
\caption{Power-error decomposition using a diagnostic multiplier fitted per household--appliance group. Values are macro-averaged over 14 groups, a different weighting from the equal-corpus main result. Factors above one indicate under-prediction.}
\label{tab:scale-decomp}
\small
\begin{tabular*}{\linewidth}{@{\extracolsep{\fill}}lrr@{}}
\toprule
Metric & All seven corpora & REFIT held out \\
\midrule
Actual MAE-on (W) & 202 $\pm$ 3 & 249 $\pm$ 16 \\
Rescaled MAE-on (W) & 163 $\pm$ 7 & 188 $\pm$ 9 \\
Scale share & 19\% & 25\% \\
Median factor: seen & 1.08 & 1.08 \\
Median factor: REFIT & 1.25 & 2.11 \\
Under-predicted groups & 93\% & 100\% \\
\bottomrule
\end{tabular*}
\end{table}

\subsection{Deployment Diagnostics}
\label{sec:results-covariates}
\label{sec:results-geometry}
\label{sec:results-deployment}
\label{sec:results-cadence}
\label{sec:results-retro}
\label{sec:results-paraphrase}

\S\ref{sec:results-zeroshot} predicted that multi-phase loads are limited by
prompt span, not exemplar count. Widening the request window to 1920\,s
confirms it: per-instant AUPRC rises 9.23 points for exactly those categories
while MAE-on stays flat (Table~\ref{tab:geometry}).
Calendar context is the complementary case of information genuinely present:
time-of-day and weekday channels add 2.71 points of event F1 and 2.39 of
AUPRC, and
the F1 gain vanishes once timestamps are shuffled, so the model reads real
usage rhythm rather than an extra channel (Table~\ref{tab:covariates}). Both
forms of context, when an appliance runs and how long, sharpen detection and
leave watt calibration untouched, consistent with
\S\ref{sec:results-scale}.

Streaming adds a third kind of context, adjacent windows, without touching
the model (Table~\ref{tab:streaming}). Pooling event scores over the current
and preceding 120\,s raises pooled event F1 from 55.63\% to 57.05\%
($+1.42$ points [+1.12, +1.78]) at the fixed threshold; REDD and UK-DALE
benefit, REFIT does not, its selected windows having no neighbor in range.
Sixty seconds of lookahead lifts it to $+2.52$ and further delay adds
nothing, so a one-minute buffer buys all of it. AUPRC
barely moves: pooling repairs a miscalibrated operating point, it does not
improve the ranking. The hurdle
factorization also exposes the limit of the trick: applying the same
averaging to the power channel flattens active peaks and costs 70.8\,W
(uniform) or 36.5\,W (Hann), so the deployed stream pools state scores and
leaves power trajectories alone. Cadence diversity helps on both axes: multi-rate copies of the same streams add 9.19 points of event F1 and
8.80 of AUPRC while removing 64\,W, strengthening the representation where
output-side averaging can only blur it.

One source of household scale does travel, and it is already in the
measurement. Pairing a request-compatible rise with its matching fall reads
the appliance's level directly off the meter as a step height: anchoring
such intervals retroactively cuts UK-DALE MAE-on from 753 to 470\,W and
lifts the amplitude ratio from 0.44 to 0.84 (Table~\ref{tab:retro}). Level
that cannot be shipped in a prompt can sometimes be recovered from the
aggregate itself. Request wording, finally, bounds how far language alone
reaches: held-out colloquial phrasings retain 92\% of canonical AUPRC for
refrigerators and 84\% for washing machines but only 35--44\% for microwave,
kettle, and dishwasher. Categories whose electrical identity a casual
sentence underdetermines are exactly those that need exemplars. Language
routes the request; signal grounds it.
Appendix~\ref{app:deployment-details} gives the full tables and a
qualitative example.

\section{Discussion and Conclusion}
\label{sec:discussion}

\paragraph{What a request can and cannot carry.}
Four ingredients make one set of weights answerable by name, and none is
dispensable (\S\ref{sec:results-ablation}): label-free pretraining builds a
power representation before any appliance is named, a frozen language
encoder turns it into an open and stably addressed output space that a
learned index does not match, exemplars supply the electrical grounding
language underdetermines, and cadence-aware tokenization over physical time
bridges meters an order of magnitude apart in rate. The same logic bounds
the interface, because a request transfers only what it can physically
carry. Exemplars serve loads whose signature fits inside one window and do
little for multi-phase cycles, where the remedy is span rather than more
examples; enrollment absorbs amplitude jitter and truncation but fails past
one mislabeled example in five. The substantive open problem is absolute
watt level: a held-out corpus is under-predicted in every
household--appliance group, and an oracle multiplier removes a quarter of
the error, leaving calibration rather than shape. This reflects
cross-household transfer rather than a cost of the interface, since \fm
still records the lowest MAE-on of any model trained here and the hurdle
factorization merely makes the residual nameable. Four request-side
parameterizations failed to close it, whereas edge-anchored step heights
recover part of it from the aggregate, so level is best measured in the
target home rather than shipped in the prompt. Our evidence covers
active-power submetering at 1--60\,s cadences in public corpora; sub-second
and reactive-power regimes lie outside it.

\paragraph{Conclusion.}
\fm makes the set of answerable appliances an argument of the forward pass
rather than a property of the weights. One frozen, lightweight
\fm model serves twelve appliance--corpus requests that would otherwise
occupy twelve networks, matches the strongest specialist on detection at the
lowest watt error we measured, and answers requests for appliances and
corpora that its supervision never described. The contribution is an
interface as much as a model: disaggregation posed as prompt-conditioned
inverse inference, a recipe that makes heterogeneous, partially labeled
corpora trainable under one parameter set, and an evaluation that separates
what a request transfers from what it cannot. For a field whose deployments
have grown by accumulating models, coverage is better grown through the
questions a service can ask than through the inventory it must maintain.

\bibliographystyle{ACM-Reference-Format}
\bibliography{references}

@article{hart1992nonintrusive,
  title={Nonintrusive appliance load monitoring},
  author={Hart, George W},
  journal={Proceedings of the IEEE},
  volume={80},
  number={12},
  pages={1870--1891},
  year={1992},
  publisher={IEEE}
}

@inproceedings{kolter2011redd,
  title={REDD: A public data set for energy disaggregation research},
  author={Kolter, J Zico and Johnson, Matthew J},
  booktitle={Workshop on data mining applications in sustainability (SIGKDD), San Diego, CA},
  volume={25},
  number={Citeseer},
  pages={59--62},
  year={2011}
}

@article{kelly2015ukdale,
  title={The UK-DALE dataset, domestic appliance-level electricity demand and whole-house demand from five UK homes},
  author={Kelly, Jack and Knottenbelt, William},
  journal={Scientific data},
  volume={2},
  number={1},
  pages={150007},
  year={2015},
  publisher={Nature Publishing Group}
}

@article{murray2017refit,
  title={An electrical load measurements dataset of United Kingdom households from a two-year longitudinal study},
  author={Murray, David and Stankovic, Lina and Stankovic, Vladimir},
  journal={Scientific data},
  volume={4},
  number={1},
  pages={160122},
  year={2017},
  publisher={Nature Publishing Group}
}

@inproceedings{beckel2014eco,
  title={The ECO data set and the performance of non-intrusive load monitoring algorithms},
  author={Beckel, Christian and Kleiminger, Wilhelm and Cicchetti, Romano and Staake, Thorsten and Santini, Silvia},
  booktitle={Proceedings of the 1st ACM conference on embedded systems for energy-efficient buildings},
  pages={80--89},
  year={2014}
}

@article{makonin2016ampds2,
  title={Ampds2: the almanac of minutely power dataset},
  author={Makonin, Stephen},
  journal={Harvard Dataverse dataset},
  pages={48},
  year={2016}
}

@inproceedings{monacchi2014greend,
  title={GREEND: An energy consumption dataset of households in Italy and Austria},
  author={Monacchi, Andrea and Egarter, Dominik and Elmenreich, Wilfried and D'Alessandro, Salvatore and Tonello, Andrea M},
  booktitle={2014 IEEE International Conference on Smart Grid Communications (SmartGridComm)},
  pages={511--516},
  year={2014},
  organization={IEEE}
}

@article{pullinger2021ideal,
  title={The IDEAL household energy dataset, electricity, gas, contextual sensor data and survey data for 255 UK homes},
  author={Pullinger, Martin and Kilgour, Jonathan and Goddard, Nigel and Berliner, Niklas and Webb, Lynda and Dzikovska, Myroslava and Lovell, Heather and Mann, Janek and Sutton, Charles and Webb, Janette and others},
  journal={Scientific Data},
  volume={8},
  number={1},
  pages={146},
  year={2021},
  publisher={Nature Publishing Group UK London}
}

@inproceedings{kolter2012fhmm,
  title={Approximate inference in additive factorial hmms with application to energy disaggregation},
  author={Kolter, J Zico and Jaakkola, Tommi},
  booktitle={Artificial intelligence and statistics},
  pages={1472--1482},
  year={2012},
  organization={PMLR}
}

@inproceedings{kelly2015neural,
  title={Neural nilm: Deep neural networks applied to energy disaggregation},
  author={Kelly, Jack and Knottenbelt, William},
  booktitle={Proceedings of the 2nd ACM international conference on embedded systems for energy-efficient built environments},
  pages={55--64},
  year={2015}
}

@inproceedings{zhang2018sequence,
  title={Sequence-to-point learning with neural networks for non-intrusive load monitoring},
  author={Zhang, Chaoyun and Zhong, Mingjun and Wang, Zongzuo and Goddard, Nigel and Sutton, Charles},
  booktitle={Proceedings of the AAAI conference on artificial intelligence},
  volume={32},
  number={1},
  year={2018}
}

@inproceedings{shin2019subtask,
  title={Subtask gated networks for non-intrusive load monitoring},
  author={Shin, Changho and Joo, Sunghwan and Yim, Jaeryun and Lee, Hyoseop and Moon, Taesup and Rhee, Wonjong},
  booktitle={Proceedings of the AAAI conference on artificial intelligence},
  volume={33},
  number={01},
  pages={1150--1157},
  year={2019}
}

@inproceedings{yue2020bert4nilm,
  title={Bert4nilm: A bidirectional transformer model for non-intrusive load monitoring},
  author={Yue, Zhenrui and Witzig, Camilo Requena and Jorde, Daniel and Jacobsen, Hans-Arno},
  booktitle={Proceedings of the 5th international workshop on non-intrusive load monitoring},
  pages={89--93},
  year={2020}
}

@inproceedings{yue2022eltransformer,
  title={Efficient localness transformer for smart sensor-based energy disaggregation},
  author={Yue, Zhenrui and Zeng, Huimin and Kou, Ziyi and Shang, Lanyu and Wang, Dong},
  booktitle={2022 18th International Conference on Distributed Computing in Sensor Systems (DCOSS)},
  pages={141--148},
  year={2022},
  organization={IEEE}
}

@article{xiong2023matnilm,
  title={MATNilm: Multi-appliance-task non-intrusive load monitoring with limited labeled data},
  author={Xiong, Jing and Hong, Tianqi and Zhao, Dongbo and Zhang, Yu},
  journal={IEEE Transactions on Industrial Informatics},
  volume={20},
  number={3},
  pages={3177--3187},
  year={2023},
  publisher={IEEE}
}

@inproceedings{he2023msdc,
  title={MSDC: Exploiting multi-state power consumption in non-intrusive load monitoring based on a dual-CNN model},
  author={He, Jialing and Liu, Jiamou and Zhang, Zijian and Chen, Yang and Liu, Yiwei and Khoussainov, Bakh and Zhu, Liehuang},
  booktitle={Proceedings of the AAAI Conference on Artificial Intelligence},
  volume={37},
  number={4},
  pages={5078--5086},
  year={2023}
}

@article{ding2024appliancefilter,
  title={ApplianceFilter: Targeted electrical appliance disaggregation with prior knowledge fusion},
  author={Ding, Dong and Li, Junhuai and Wang, Huaijun and Wang, Kan and Feng, Jie and Xiao, Ming},
  journal={Applied Energy},
  volume={365},
  pages={123157},
  year={2024},
  publisher={Elsevier}
}

@inproceedings{toirov2025continual,
  title={Non-Intrusive Load Monitoring Based on Image Load Signatures and Continual Learning},
  author={Toirov, Olimjon and Yu, Wei},
  booktitle={Proceedings of the 2025 2nd International Conference on Digital Society and Artificial Intelligence},
  pages={894--899},
  year={2025}
}

@article{liu2026lab,
  title={Latent abstraction bridge transformer for generalizable nonintrusive load monitoring},
  author={Liu, Yuan and Kong, Zhengmin and Huang, Tao and Yang, Yang and Song, Chenjie and Han, Qing-Long and Huang, Boyang},
  journal={Scientific Reports},
  year={2026},
  publisher={Nature Publishing Group UK London}
}

@inproceedings{xue2025prompting,
  title={Prompting large language models for training-free non-intrusive load monitoring},
  author={Xue, Junyu and Wang, Xudong and He, Xiaoling and Liu, Shicheng and Wang, Yi and Tang, Guoming},
  booktitle={Proceedings of the 12th ACM International Conference on Systems for Energy-Efficient Buildings, Cities, and Transportation},
  pages={34--44},
  year={2025}
}

@inproceedings{das2024decoder,
  author={Das, Abhimanyu and Kong, Weihao and Sen, Rajat and Zhou, Yichen},
  title={A Decoder-Only Foundation Model for Time-Series Forecasting},
  booktitle={Proceedings of the 41st International Conference on Machine Learning},
  year={2024}
}

@article{goswami2024moment,
  title={Moment: A family of open time-series foundation models},
  author={Goswami, Mononito and Szafer, Konrad and Choudhry, Arjun and Cai, Yifu and Li, Shuo and Dubrawski, Artur},
  journal={arXiv preprint arXiv:2402.03885},
  year={2024}
}

@inproceedings{liu2025moiraimoe,
  author={Liu, Xu and Liu, Juncheng and Woo, Gerald and Aksu, Taha and Liang, Yuxuan and Zimmermann, Roger and Liu, Chenghao and Li, Junnan and Savarese, Silvio and Xiong, Caiming and Sahoo, Doyen},
  title={{Moirai-MoE}: Empowering Time Series Foundation Models with Sparse Mixture of Experts},
  booktitle={Proceedings of the 42nd International Conference on Machine Learning},
  volume={267},
  pages={38940--38962},
  year={2025}
}

@article{ansari2025chronos2,
  author={Ansari, Abdul Fatir and Shchur, Oleksandr and K{\"u}ken, Jaris and Auer, Andreas and Han, Boran and Mercado, Pedro and Rangapuram, Syama Sundar and Shen, Huibin and Stella, Lorenzo and Zhang, Xiyuan and others},
  title={{Chronos-2}: From Univariate to Universal Forecasting},
  journal={arXiv preprint arXiv:2510.15821},
  year={2025}
}

@article{liu2024timer,
  author={Liu, Yong and Zhang, Haoran and Li, Chenyu and Huang, Xiangdong and Wang, Jianmin and Long, Mingsheng},
  title={{Timer}: Generative Pre-Trained Transformers Are Large Time Series Models},
  journal={arXiv preprint arXiv:2402.02368},
  year={2024}
}

@inproceedings{shi2024timemoe,
title={Time-MoE: Billion-Scale Time Series Foundation Models with Mixture of Experts},
author={Xiaoming Shi and Shiyu Wang and Yuqi Nie and Dianqi Li and Zhou Ye and Qingsong Wen and Ming Jin},
booktitle={The Thirteenth International Conference on Learning Representations},
year={2025},
url={https://openreview.net/forum?id=e1wDDFmlVu}
}

@inproceedings{radford2021clip,
  author={Radford, Alec and Kim, Jong Wook and Hallacy, Chris and Ramesh, Aditya and Goh, Gabriel and Agarwal, Sandhini and Sastry, Girish and Askell, Amanda and Mishkin, Pamela and Clark, Jack and others},
  title={Learning Transferable Visual Models from Natural Language Supervision},
  booktitle={Proceedings of the International Conference on Machine Learning},
  pages={8748--8763},
  year={2021}
}

@inproceedings{wang2021tent,
  author={Wang, Dequan and Shelhamer, Evan and Liu, Shaoteng and Olshausen, Bruno and Darrell, Trevor},
  title={Tent: Fully Test-Time Adaptation by Entropy Minimization},
  booktitle={International Conference on Learning Representations},
  year={2021}
}

@article{zhang2025qwen3embedding,
  author={Zhang, Yanzhao and Li, Mingxin and Long, Dingkun and Zhang, Xin and Lin, Huan and Yang, Baosong and Xie, Pengjun and Yang, An and Liu, Dayiheng and Lin, Junyang and Huang, Fei and Zhou, Jingren},
  title={{Qwen3} Embedding: Advancing Text Embedding and Reranking Through Foundation Models},
  journal={arXiv preprint arXiv:2506.05176},
  year={2025}
}

@article{garciamarrero2026refquery,
  author={Garcia-Marrero, L. E. and Petrone, G. and Monmasson, E.},
  title={Lightweight and Scalable Transfer Learning Framework for Load Disaggregation},
  journal={arXiv preprint arXiv:2603.04998},
  year={2026}
}

@article{delcroix2022soundbeam,
  author={Delcroix, Marc and V{\'a}zquez, Jorge Bennasar and Ochiai, Tsubasa and Kinoshita, Keisuke and Ohishi, Yasunori and Araki, Shoko},
  title={{SoundBeam}: Target Sound Extraction Conditioned on Sound-Class Labels and Enrollment Clues for Increased Performance and Continuous Learning},
  journal={arXiv preprint arXiv:2204.03895},
  year={2022}
}

@article{liu2023audiosep,
  author={Liu, Xubo and Kong, Qiuqiang and Zhao, Yan and Liu, Haohe and Yuan, Yi and Liu, Yuzhuo and Xia, Rui and Wang, Yuxuan and Plumbley, Mark D. and Wang, Wenwu},
  title={Separate Anything You Describe},
  journal={arXiv preprint arXiv:2308.05037},
  year={2023}
}

@inproceedings{cai2025detectanysound,
author = {Cai, Pengfei and Song, Yan and Gu, Qing and Jiang, Nan and Song, Haoyu and McLoughlin, Ian},
title = {Detect Any Sound: Open-Vocabulary Sound Event Detection with Multi-Modal Queries},
year = {2025},
isbn = {9798400720352},
publisher = {Association for Computing Machinery},
address = {New York, NY, USA},
url = {https://doi.org/10.1145/3746027.3755574},
booktitle = {Proceedings of the 33rd ACM International Conference on Multimedia},
pages = {582–591},
numpages = {10},
location = {Dublin, Ireland},
series = {MM '25}
}

@INPROCEEDINGS{hai2025flexsed,
  author={Hai, Jiarui and Wang, Helin and Guo, Weizhe and Elhilali, Mounya},
  booktitle={2025 IEEE Workshop on Applications of Signal Processing to Audio and Acoustics (WASPAA)}, 
  title={FlexSED: Towards Open-Vocabulary Sound Event Detection}, 
  year={2025},
  volume={},
  number={},
  pages={1-5},
  doi={10.1109/WASPAA66052.2025.11230975}
  }

@article{li2024tlinet,
  author={Li, Danyang and Cai, Mingyu and Vasile, Cristian-Ioan and Tron, Roberto},
  title={{TLINet}: Differentiable Neural Network Temporal Logic Inference},
  journal={arXiv preprint arXiv:2405.06670},
  year={2024}
}

@article{gruver2023llmtime,
  author={Gruver, Nate and Finzi, Marc and Qiu, Shikai and Wilson, Andrew Gordon},
  title={Large Language Models Are Zero-Shot Time Series Forecasters},
  journal={Advances in Neural Information Processing Systems},
  volume={36},
  year={2023}
}

@inproceedings{jin2023timellm,
 author = {Jin, Ming and Wang, Shiyu and Ma, Lintao and Chu, Zhixuan and Zhang, James and Shi, Xiaoming and Chen, Pin-Yu and Liang, Yuxuan and Li, Yuan-Fang and Pan, Shirui and Wen, Qingsong},
 booktitle = {International Conference on Learning Representations},
 editor = {B. Kim and Y. Yue and S. Chaudhuri and K. Fragkiadaki and M. Khan and Y. Sun},
 pages = {23857--23880},
 title = {Time-LLM: Time Series Forecasting by Reprogramming Large Language Models},
 url = {https://proceedings.iclr.cc/paper_files/paper/2024/file/680b2a8135b9c71278a09cafb605869e-Paper-Conference.pdf},
 volume = {2024},
 year = {2024}
}

@inproceedings{petralia2025nilmformer,
  author={Petralia, Adrien and Charpentier, Philippe and Kadhi, Youssef and Palpanas, Themis},
  title={{NILMFormer}: Non-Intrusive Load Monitoring that Accounts for Non-Stationarity},
  booktitle={Proceedings of the 31st ACM SIGKDD Conference on Knowledge Discovery and Data Mining V.2},
  pages={4761--4772},
  year={2025},
  publisher={Association for Computing Machinery},
  doi={10.1145/3711896.3737251}
}

@ARTICLE{chen2026sceneaware,
  author={Chen, Haiwen and Chen, Jianning and Chai, Yuanyuan and Guo, Wei and Jia, Chen and Yang, Bin and Xin, Zhen},
  journal={IEEE Transactions on Smart Grid}, 
  title={Scene-Aware Non-Intrusive Load Monitoring Using Large Language Models}, 
  year={2026},
  volume={17},
  number={1},
  pages={874-876},
  doi={10.1109/TSG.2025.3632609}}

@inproceedings{chen2016xgboost,
  title={{XGBoost}: A Scalable Tree Boosting System},
  author={Chen, Tianqi and Guestrin, Carlos},
  booktitle={Proceedings of the 22nd ACM SIGKDD International Conference on Knowledge Discovery and Data Mining},
  pages={785--794},
  year={2016}
}

@inproceedings{wang2022evsense,
  title={Evsense: A robust and scalable approach to non-intrusive ev charging detection},
  author={Wang, Xudong and Tang, Guoming and Wang, Yi and Keshav, Srinivasan and Zhang, Yu},
  booktitle={Proceedings of the Thirteenth ACM International Conference on Future Energy Systems},
  pages={307--319},
  year={2022}
}

@inproceedings{wang2026energy,
  title={Energy Injection Identification enabled Disaggregation with Deep Multi-Task Learning},
  author={Wang, Xudong and Tang, Guoming and Xue, Junyu and Keshav, Srinivasan and Li, Tongxin and Ding, Chris},
  booktitle={Proceedings of the 17th ACM International Conference on Future and Sustainable Energy Systems},
  pages={89--119},
  year={2026}
}

\appendix

\section{Additional Deployment Evidence}
\label{app:deployment-details}

The following tables and qualitative example provide the detailed evidence behind the deployment diagnostics summarized in \S\ref{sec:results-deployment}.

\begin{table*}[!tbp]
\centering
\caption{Covariates and watt-head parameterizations on the frozen test set, reported as paired differences from the deployed operator. Shuffling timestamps preserves their marginal distribution while removing their association with each window. Calendar channels improve both detection metrics; static physical covariates and the tested watt heads provide no consistent MAE-on gain.}
\label{tab:covariates}
\small
\begin{tabular*}{\linewidth}{@{\extracolsep{\fill}}lrrrr@{}}
\toprule
Arm & seeds & $\Delta$ event F1 (pts) & $\Delta$ AUPRC (pts) & $\Delta$ MAE-on (W) \\
\midrule
calendar time & 5 & $+2.71$ [$+1.01$, $+4.34$] & $+2.39$ [$+0.79$, $+4.72$] & $-3.2$ [$-30.0$, $+17.2$] \\
\quad{}control: timestamps shuffled & 5 & $-0.36$ [$-1.80$, $+0.70$] & $+1.15$ [$-0.52$, $+2.82$] & $-13.7$ [$-50.1$, $+12.6$] \\
window-relative normalization & 5 & $-0.96$ [$-2.21$, $+0.13$] & $+4.67$ [$+3.76$, $+5.26$] & $-3.7$ [$-34.1$, $+19.3$] \\
local dynamics & 3 & $+0.05$ [$-2.03$, $+2.89$] & $+1.27$ [$+0.08$, $+3.02$] & $+5.5$ [$-12.9$, $+23.7$] \\
all channels & 5 & $+1.49$ [$-0.31$, $+3.02$] & $+3.52$ [$+1.64$, $+5.16$] & $+6.4$ [$-29.0$, $+28.6$] \\
static physical covariates & 3 & $+0.21$ [$-3.83$, $+2.81$] & $-2.47$ [$-12.98$, $+3.26$] & $+49.6$ [$-3.5$, $+132.8$] \\
aggregate-anchored watt head & 5 & $-0.32$ [$-2.04$, $+1.39$] & $+2.55$ [$-0.23$, $+5.34$] & $-8.8$ [$-40.4$, $+15.9$] \\
aggregate-share watt head & 5 & $+1.19$ [$-0.48$, $+2.25$] & $+3.91$ [$+2.79$, $+4.88$] & $-0.0$ [$-32.9$, $+20.6$] \\
\bottomrule
\end{tabular*}
\end{table*}

\begin{table*}[!tbp]
\begin{minipage}[t]{0.48\textwidth}

\centering
\caption{Appliance-state predictability from wall-clock time. A prior over hour and weekday is fitted on training data and scored on validation data; lift is its AUPRC divided by the base rate. Human-started loads show stronger schedule structure than thermostatic loads. The table lists the top 16 by lift.}
\label{tab:calendar}
\small
\setlength{\tabcolsep}{2pt}
\begin{tabular*}{\linewidth}{@{\extracolsep{\fill}}llrrrr@{}}
\toprule
Corpus & Appliance & Base (\%) & AUPRC (\%) & Lift & Peak \\
\midrule
ECO & microwave & 0.8 & 11.6 & 14.8$\times$ & 11:00 \\
GREEND & dishwasher & 0.2 & 1.9 & 8.6$\times$ & 18:00 \\
AMPds2 & oven & 2.0 & 8.1 & 4.1$\times$ & 00:00 \\
REFIT & microwave & 0.2 & 0.8 & 3.9$\times$ & 18:00 \\
UK-DALE & dishwasher & 5.8 & 21.4 & 3.7$\times$ & 19:00 \\
REDD & microwave & 0.4 & 1.1 & 2.9$\times$ & 17:00 \\
IDEAL & toaster & 1.4 & 3.3 & 2.4$\times$ & 07:00 \\
IDEAL & microwave & 3.4 & 7.9 & 2.3$\times$ & 16:00 \\
UK-DALE & microwave & 0.9 & 1.9 & 2.1$\times$ & 11:00 \\
IDEAL & dishwasher & 6.5 & 13.0 & 2.0$\times$ & 18:00 \\
REFIT & washing mach. & 2.8 & 5.1 & 1.8$\times$ & 10:00 \\
ECO & kettle & 4.1 & 6.5 & 1.6$\times$ & 07:00 \\
IDEAL & washing mach. & 10.0 & 14.8 & 1.5$\times$ & 13:00 \\
GREEND & fridge & 2.4 & 3.5 & 1.5$\times$ & 20:00 \\
UK-DALE & kettle & 0.8 & 1.1 & 1.5$\times$ & 20:00 \\
REFIT & kettle & 0.9 & 1.2 & 1.3$\times$ & 20:00 \\
\bottomrule
\end{tabular*}
\end{minipage}\hfill
\begin{minipage}[t]{0.48\textwidth}

\centering
\caption{Activation duration against the physical span of a request. ``Fits'' is the fraction of complete activations shorter than the window and therefore visible end to end. Multi-phase loads have the lowest coverage under the default span.}
\label{tab:window-span}
\small
\setlength{\tabcolsep}{2pt}
\begin{tabular*}{\linewidth}{@{\extracolsep{\fill}}llrrrr@{}}
\toprule
Corpus & Appliance & Window (s) & Median (s) & Fits (\%) & $n$ \\
\midrule
REDD & fridge & 480 & 1220 & 3 & 582 \\
REDD & dishwasher & 480 & 110 & 86 & 180 \\
REDD & microwave & 480 & 60 & 100 & 246 \\
UK-DALE & dishwasher & 480 & 886 & 35 & 69 \\
UK-DALE & fridge & 480 & 879 & 0 & 708 \\
UK-DALE & kettle & 480 & 166 & 100 & 128 \\
UK-DALE & toaster & 480 & 123 & 100 & 41 \\
UK-DALE & microwave & 480 & 36 & 100 & 169 \\
REFIT & dishwasher & 960 & 4112 & 31 & 263 \\
REFIT & washing mach. & 960 & 3944 & 11 & 250 \\
REFIT & kettle & 960 & 104 & 100 & 2349 \\
REFIT & microwave & 960 & 80 & 100 & 1410 \\
\bottomrule
\end{tabular*}
\end{minipage}
\end{table*}

\begin{table*}[!tbp]
\centering
\caption{Request geometry scored per instant on the reconstructed timeline. Physical span and token span vary independently on the same recordings. The reference row gives absolute values and the remaining rows give paired differences. A 1920\,s span gives the largest gain for multi-phase categories; extending to two hours reduces the number of independent training windows and reverses the overall gain. MAE-on remains stable across the 480--1920\,s settings.}
\label{tab:geometry}
\small
\begin{tabular*}{\linewidth}{@{\extracolsep{\fill}}rrrrrr@{}}
\toprule
Span (s) & Token (s) & ts-AUPRC (\%) & ts-AUPRC, multi-phase (\%) & MAE-on (W) & seeds \\
\midrule
480 & 4 & 74.57 & 84.16 & 219 & \multicolumn{1}{c}{reference} \\
1920 & 16 & $+2.49$ [$+0.62$, $+6.12$] & $+9.23$ [$+7.81$, $+10.74$] & $+20$ [$-64$, $+126$] & 3 \\
1920 & 4 & $+1.53$ [$-0.77$, $+5.51$] & $+2.76$ [$-3.75$, $+8.63$] & $+27$ [$+5$, $+42$] & 3 \\
7200 & 16 & $-6.07$ [$-8.66$, $-0.95$] & $-6.24$ [$-17.83$, $+3.58$] & $+158$ [$+42$, $+234$] & 3 \\
7200 & 60 & $-5.39$ [$-8.49$, $-0.84$] & $+2.37$ [$-3.02$, $+5.42$] & $+161$ [$+82$, $+260$] & 3 \\
\bottomrule
\end{tabular*}
\end{table*}

\begin{table*}[!tbp]
\centering
\caption{Streaming inference with the final \fm model over five seeds. State rows use the fixed 0.5 threshold and report paired changes from single-window inference. REDD and UK-DALE have 7.98-fold realized coverage; the selected REFIT windows have no neighbor within 120\,s. Power rows modify power trajectories while preserving event scores.}
\label{tab:streaming}
\small
\setlength{\tabcolsep}{3pt}
\begin{tabular*}{\linewidth}{@{\extracolsep{\fill}}llrrrrrr@{}}
\toprule
Channel & Context / rule & Delay & \multicolumn{4}{c}{$\Delta$ event F1 (pts)} & Pooled $\Delta$ \\
\cmidrule(lr){4-7}
& & & REDD & UK-DALE & REFIT & pooled & AUPRC (pts) / MAE-on \\
\midrule
State & past 120\,s & 0\,s & +2.78 & +1.49 & +0.00 & +1.42 & $-0.41$ / $+0.0$\,W \\
State & past 120\,s + future 60\,s & 60\,s & +4.48 & +3.08 & +0.00 & +2.52 & $-0.14$ / $+0.0$\,W \\
State & past 120\,s + future 120\,s & 120\,s & +4.48 & +3.09 & +0.00 & +2.52 & $-0.12$ / $+0.0$\,W \\
\midrule
Power & uniform average & n/a & +0.00 & +0.00 & +0.00 & +0.00 & $+70.8$\,W \\
Power & Hann average & n/a & +0.00 & +0.00 & +0.00 & +0.00 & $+36.5$\,W \\
Power & nearest center & n/a & +0.00 & +0.00 & +0.00 & +0.00 & $+0.0$\,W$^{\dagger}$ \\
\bottomrule
\end{tabular*}
\par\vspace{2pt}\raggedright\small
State pooled $\Delta$F1 95\% CIs (points): causal [+1.12, +1.78], 60\,s
delay [+2.00, +3.36], 120\,s delay [+2.00, +3.39]. Power pooled
$\Delta$MAE-on CIs: uniform [+60.7, +79.9] and Hann [+32.4, +40.2]\,W.
$^{\dagger}$Nearest-center selection increases the full-trajectory energy ratio by 0.287.
\end{table*}

\begin{table*}[!tbp]
\centering
\caption{Retroactive edge anchoring over five seeds. A switch-off paired with an earlier switch-on defines an interval whose watts are set from the measured step height. ``Attributable'' covers categories described by a separable, uninterrupted power plateau. Other categories retain their original outputs, diluting the effect in the ``all requested'' rows. The rule updates the watt channel, so detection remains unchanged up to numerical noise.}
\label{tab:retro}
\small
\begin{tabular*}{\linewidth}{@{\extracolsep{\fill}}llrrrrrrr@{}}
\toprule
& & \multicolumn{3}{c}{MAE-on (W)} & \multicolumn{3}{c}{amplitude ratio} & ts-AUPRC (pts) \\
\cmidrule(lr){3-5}\cmidrule(lr){6-8}
Scope & Corpus & base & revised & $\Delta$ & base & revised & $\Delta$ & $\Delta$ \\
\midrule
attributable & REDD & 730 & 709 & $-20$ [$-91$, $+41$] & 0.543 & 0.670 & $+0.127$ [$+0.086$, $+0.175$] & $+0.00$ \\
attributable & UK-DALE & 753 & 470 & $-283$ [$-338$, $-217$] & 0.440 & 0.840 & $+0.400$ [$+0.358$, $+0.426$] & $+0.00$ \\
attributable & REFIT & 824 & 703 & $-121$ [$-140$, $-107$] & 0.558 & 0.721 & $+0.163$ [$+0.154$, $+0.171$] & $+0.00$ \\
all requested & REDD & 282 & 275 & $-7$ [$-30$, $+14$] & 0.755 & 0.798 & $+0.042$ [$+0.029$, $+0.058$] & $+0.00$ \\
all requested & UK-DALE & 473 & 303 & $-170$ [$-203$, $-130$] & 0.612 & 0.852 & $+0.240$ [$+0.215$, $+0.255$] & $+0.00$ \\
all requested & REFIT & 566 & 506 & $-61$ [$-70$, $-54$] & 0.650 & 0.731 & $+0.081$ [$+0.077$, $+0.085$] & $+0.00$ \\
\bottomrule
\end{tabular*}
\end{table*}

\begin{table*}[!tbp]
\centering
\small
\caption{Prompt-phrasing robustness over five seeds. ``Canonical'' is the structured appliance card; ``user phrasings'' are held-out colloquial requests for the same appliance. Retention is mean user-phrasing AUPRC divided by canonical AUPRC. Categories with distinctive electrical signatures retain the most accuracy.}
\label{tab:paraphrase}
\setlength{\tabcolsep}{2pt}
\begin{tabular*}{\linewidth}{@{\extracolsep{\fill}}lrrrrrr@{}}
\toprule
AUPRC (\%) & Fridge & Microwave & Dishwasher & Kettle & Toaster & Washing machine \\
\midrule
Canonical & 79.5 & 50.0 & 72.9 & 75.6 & 20.4 & 41.1 \\
User mean & 72.9 & 17.4 & 32.3 & 27.0 & 10.9 & 34.5 \\
User worst & 66.7 & 9.1 & 13.3 & 4.3 & 4.5 & 28.4 \\
Retention & 92\% & 35\% & 44\% & 36\% & 54\% & 84\% \\
\bottomrule
\end{tabular*}
\end{table*}

\section{Additional Controls}
\label{app:controls}

Table~\ref{tab:controls} lists the exact values behind Fig.~\ref{fig:controls}. The released evaluation also includes three inference-time interventions: canonical cards versus held-out user phrasings, exemplar-count sweeps for $K \in \{1,3,5,10\}$, and same-corpus versus cross-corpus exemplars under dataset exclusion.

\begin{table*}[!tbp]
\centering
\caption{Causal controls at inference (five seeds).  Text controls apply to the full model over all requests; exemplar controls apply to the held-out microwave under the unseen-appliance protocol, in exemplar-only mode.  These are the values plotted in Fig.~\ref{fig:controls}.}
\label{tab:controls}
\small
\begin{tabular*}{0.62\linewidth}{@{\extracolsep{\fill}}llr@{}}
\toprule
Modality & Intervention & AUPRC (\%) \\
\midrule
\multirow{4}{*}{Text} & canonical card & 62.5 $\pm$ 2.2 \\
 & matched confuser & 27.4 $\pm$ 0.7 \\
 & deranged permutation & 13.8 $\pm$ 1.9 \\
 & zero signal & 7.8 $\pm$ 0.0 \\
\midrule
\multirow{2}{*}{Exemplar} & correct category ($K{=}5$) & 17.1 $\pm$ 2.3 \\
 & wrong category ($K{=}5$) & 1.3 $\pm$ 0.3 \\
\bottomrule
\end{tabular*}
\end{table*}

\section{Reproducibility Details}
\label{app:repro}

\paragraph{Artifacts.}
All processed windows, splits, event indices, thresholds, and test manifests are fixed before training and recorded with 256-bit Secure Hash Algorithm (SHA-256) digests. The artifact contains download and preprocessing instructions, array schemas, split and event manifests, and integrity hashes. Authorized copies of the seven public corpora are required to reconstruct the arrays used by the training and evaluation scripts.

\paragraph{Model.}
Trunk: patch size 4, width 192, 6 pre-normalization transformer layers, 4 heads, FFN width 768, dropout 0.1; physical-time sinusoidal encodings with periods 10\,s--7200\,s; convolutional upsampler (kernels 5, 3). Prompt encoder: text MLP $1024{\to}384{\to}192$, exemplar MLP $192{\to}192{\to}192$, count embedding for $K \le 16$, mode embedding over \{none, text, exemplar, both\}. Decoder: 4-head cross-attention, additive fusion at width 96, linear hurdle heads, $\sigma$ clipped to $[0.05, 1.5]$, learned-temperature log-sum-exp (LSE) window pooling. Total 3{,}705{,}707 parameters.

\paragraph{Optimization.}
Pretraining: AdamW, one-cycle $3{\times}10^{-4}$, batch 256, 20 epochs, contiguous span masking at ratio 0.4 reconstructing the raw-kilowatt target under a validity-weighted Huber loss.  Alignment: AdamW, one-cycle $3{\times}10^{-4}$, weight decay 0.01, batch 128, 60 epochs $\times$ 400 steps, gradient clip 1.0, bf16 autocast.  Loss weights (Eq.~\ref{eq:loss}) were fixed on development data before any held-out split was scored; no hyperparameter of this paper was tuned on any test split.

\paragraph{Text cards.}
The full card bank (canonical card plus 6--7 paraphrases per category), the confuser mapping, and the permutation derangement are released verbatim; cards were written from generic appliance knowledge before any model in this paper was trained and never edited afterwards.

\paragraph{Exemplar library.}
Exemplars are training-split windows containing one complete activation of the category (event boundaries from the fixed event indices), with peak power above the category's frozen on-threshold.  Library sizes: fridge 13{,}279; washing machine 47{,}500; tumble dryer 18{,}094; microwave 5{,}497; dishwasher 4{,}544; kettle 3{,}700; oven 2{,}413; toaster 750.  Draws during training exclude the batch's source; draws at evaluation follow the protocol of each experiment and are repeated over five random sets with the spread reported.

\section{Dataset Statistics and Splits}
\label{app:dataset-stats}

Table~\ref{tab:corpus-stats} reports window counts and house assignments, and Table~\ref{tab:appliance-stats} gives each appliance's threshold, event count, positive-window rate, and typical on-state peak power. REFIT uses house-disjoint splits. REDD and UK-DALE use chronological 70/10/20 splits with purge gaps at the boundaries. Positive-window rates include windows where the appliance is submetered, as indicated by its observation mask. Activations are sparse and heavy-tailed: most appliances occupy a small fraction of windows, and peak powers span two orders of magnitude. These properties motivate event-balanced training and separate detection and scale heads.

\begin{table*}[!tbp]
\centering
\small
\caption{Per-corpus splits and window counts under the fixed splits. Houses are disjoint across splits for REFIT; REDD and UK-DALE use strict chronological within-house splits with purge gaps; the four auxiliary corpora supply supervision and pretraining but carry no frozen test split. All windows are non-overlapping at the listed stride.}
\label{tab:corpus-stats}
\setlength{\tabcolsep}{1pt}
\begin{tabular*}{\linewidth}{@{\extracolsep{\fill}}lrrrrrl@{}}
\toprule
Corpus & Cad. & Win & Train & Val & Test & Houses (tr/va/te) \\
\midrule
REDD & 1\,s & 480 & 25{,}844 & 3{,}577 & 7{,}332 & 1/2;1/2;1/2 \\
UKDALE & 1\,s & 480 & 23{,}147 & 3{,}203 & 6{,}580 & 1/2;1/2;1/2 \\
REFIT & 8\,s & 120 & 10{,}000 & 5{,}000 & 5{,}000 & 3/5/6/9;2/11;13/20 \\
ECO & 8\,s & 120 & 6{,}822 & 3{,}565 & 0 & 1/3/4/6;2/5;-- \\
IDEAL & 8\,s & 120 & 33{,}532 & 3{,}993 & 0 & 25 houses;105/175/311/328;-- \\
GREEND & 8\,s & 120 & 5{,}388 & 3{,}589 & 0 & 1/4/5;2/6;-- \\
AMPDS2 & 60\,s & 32 & 4{,}999 & 5{,}000 & 0 & 1;1;-- \\
\bottomrule
\end{tabular*}
\end{table*}
\begin{table*}[!tbp]
\centering
\small
\caption{Per-appliance statistics under the fixed splits: on/off threshold, number of
complete activation events, positive-window rate (fraction of observed windows in which
the appliance is active), and median on-state peak power.  Only appliances that are
actually submetered in a corpus are listed.}
\label{tab:appliance-stats}
\setlength{\tabcolsep}{2pt}
\begin{minipage}{0.495\linewidth}
\centering
\begin{tabular*}{\linewidth}{@{\extracolsep{\fill}}llrrrr@{}}
\toprule
Corpus & Appliance & Thr.\,(W) & Events & Pos.\ (\%) & Peak\,(W) \\
\midrule
REDD & fridge & 50 & 582 & 34.9 & 188 \\
REDD & microwave & 200 & 246 & 0.9 & 1575 \\
REDD & dishwasher & 50 & 180 & 2.4 & 1107 \\
UKDALE & kettle & 200 & 128 & 0.9 & 2979 \\
UKDALE & microwave & 200 & 169 & 0.5 & 1339 \\
UKDALE & fridge & 50 & 708 & 39.5 & 95 \\
UKDALE & dishwasher & 50 & 69 & 3.2 & 2029 \\
UKDALE & toaster & 200 & 41 & 0.3 & 1579 \\
REFIT & microwave & 200 & 1{,}410 & 0.5 & 1454 \\
REFIT & dishwasher & 50 & 263 & 3.0 & 2185 \\
REFIT & kettle & 1000 & 2{,}349 & 0.7 & 2704 \\
REFIT & washing\_machine & 20 & 250 & 3.1 & 500 \\
ECO & fridge & 50 & 8{,}257 & 35.6 & 129 \\
ECO & freezer & 50 & 2{,}138 & 41.2 & 213 \\
ECO & microwave & 200 & 307 & 3.3 & 1537 \\
ECO & dishwasher & 50 & 45 & 4.3 & 2182 \\
ECO & kettle & 1000 & 161 & 2.4 & 1854 \\
\bottomrule
\end{tabular*}
\end{minipage}\hfill
\begin{minipage}{0.495\linewidth}
\centering
\begin{tabular*}{\linewidth}{@{\extracolsep{\fill}}llrrrr@{}}
\toprule
Corpus & Appliance & Thr.\,(W) & Events & Pos.\ rate & Peak\,(W) \\
\midrule
ECO & washing\_machine & 20 & 795 & 0.153 & 443 \\
IDEAL & fridge & 50 & 14{,}215 & 0.443 & 109 \\
IDEAL & microwave & 200 & 855 & 0.018 & 1532 \\
IDEAL & dishwasher & 50 & 960 & 0.056 & 2232 \\
IDEAL & kettle & 1000 & 1{,}542 & 0.040 & 2867 \\
IDEAL & toaster & 800 & 353 & 0.021 & 1014 \\
IDEAL & washing\_machine & 20 & 14{,}873 & 0.099 & 420 \\
IDEAL & tumble\_dryer & 50 & 10{,}199 & 0.262 & 314 \\
IDEAL & electric\_heater & 200 & 11 & 0.286 & 1410 \\
GREEND & fridge & 50 & 1{,}197 & 0.243 & 90 \\
GREEND & freezer & 50 & 87 & 0.084 & 92 \\
GREEND & dishwasher & 50 & 819 & 0.079 & 313 \\
GREEND & kettle & 1000 & 24 & 0.005 & 1781 \\
GREEND & washing\_machine & 20 & 3{,}196 & 0.164 & 171 \\
AMPDS2 & fridge & 50 & 15{,}185 & 0.514 & 148 \\
AMPDS2 & oven & 200 & 460 & 0.032 & 3478 \\
\bottomrule
\end{tabular*}
\end{minipage}
\end{table*}

\section{Per-Dataset and Per-Appliance Results}
\label{app:per-dataset}

\begin{figure*}[!tbp]
\centering
\includegraphics[width=\linewidth]{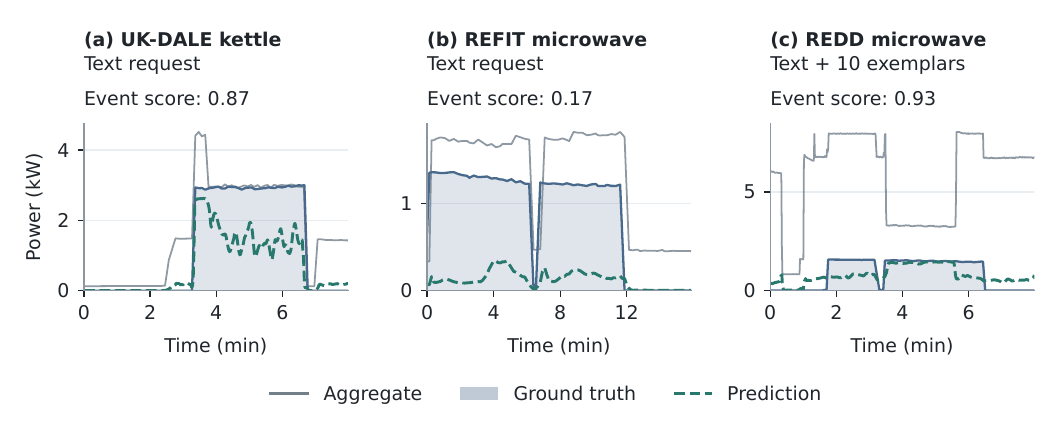}
\caption{Examples of requested disaggregation. Gray lines show aggregate power, blue traces and shading show the target, and dashed teal lines show predicted power, all in kW. \textbf{(a)}~A UK-DALE kettle requested by text. \textbf{(b)}~A REFIT microwave with a low event score and underestimated amplitude. \textbf{(c)}~A microwave excluded from supervision and pretraining submeter channels, requested with text and ten exemplars. Each panel reports its saved window event score. Table~\ref{tab:loao} summarizes category transfer across test windows.}
\Description{Three time-series panels compare aggregate, target, predicted power, and window event score for a supervised kettle, a REFIT microwave, and a held-out microwave prompted with ten exemplars.}
\label{fig:qualitative}
\end{figure*}

Table~\ref{tab:per-appliance} expands the headline comparison to individual
appliances. It places one shared \fm beside BERT4NILM, the strongest per-appliance
specialist, which trains a separate network for every row. Two patterns emerge. \fm has
lower MAE-on for most appliances, including 470\,W versus 1039\,W for the REFIT
microwave, consistent with its aggregate watt-error advantage. Event F1 is mixed:
BERT4NILM is stronger on several UK-DALE categories, and \fm is stronger on the sparse
REFIT microwave and kettle. Figure~\ref{fig:qualitative} shows representative windows for a supervised
request, a hard REFIT request, and a held-out category. Per-house results and further
operating-point diagnostics, including false power, energy ratio, and missed-event rate,
accompany the artifact as JSON.

\begin{table*}[!tbp]
\begin{minipage}[t]{0.48\textwidth}

\centering
\small
\caption{Inference efficiency (single 480-sample window, mean of 50 runs; RTX 5090 for GPU, 4-thread CPU).  \fm serves any appliance request with one network; specialists need one network per appliance, so their deployed footprint is the per-model figure times the number of appliances.  \fm's per-request latency is comparable to a single transformer specialist while its total on-meter footprint is the smallest of the neural models.}
\label{tab:efficiency}
\setlength{\tabcolsep}{4pt}
\begin{tabular*}{\linewidth}{@{\extracolsep{\fill}}lrrrl@{}}
\toprule
Model & Params & CPU\,(ms) & GPU\,(ms) & Serves \\
\midrule
Seq2Point & 24.6\,M & 1.7 & 0.3 & 1 appliance \\
BERT4NILM & 0.61\,M & 4.3 & 0.7 & 1 appliance \\
NILMFormer & 0.60\,M & 4.4 & 0.9 & 1 appliance \\
\midrule
\fm & 3.7\,M & 5.5 & 2.3 & any request \\
\bottomrule
\end{tabular*}
\end{minipage}\hfill
\begin{minipage}[t]{0.48\textwidth}

\centering
\small
\caption{Noisy-enrollment robustness for the held-out microwave ($K{=}5$, five seeds), as a percentage of clean-enrollment AUPRC.  Amplitude jitter multiplies each exemplar sample by $1{+}\mathcal{N}(0,\ell)$; truncation zeros a fraction $\ell$ of the activation; mislabeling replaces a fraction $\ell$ of exemplars with a wrong category.  Enrollment survives amplitude and truncation noise but is bounded by label quality.}
\label{tab:noisy}
\setlength{\tabcolsep}{5pt}
\begin{tabular*}{\linewidth}{@{\extracolsep{\fill}}lrrr@{}}
\toprule
Corruption & light & moderate & heavy \\
\midrule
Amplitude jitter ($\ell{=}0.1/0.2/0.4$) & 65\% & 69\% & 61\% \\
Truncation ($\ell{=}0.2/0.4/0.6$) & 116\% & 117\% & 112\% \\
Mislabeling ($\ell{=}0.2/0.4/0.6$) & 84\% & 12\% & 8\% \\
\bottomrule
\end{tabular*}
\end{minipage}
\end{table*}

\begin{table*}[!tbp]
\centering
\small
\caption{Per-corpus, per-appliance results (five-seed means): one shared \fm with a
text prompt versus BERT4NILM, the strongest per-appliance specialist with a dedicated
model for each row. Detection is mixed across categories; \fm has lower MAE-on for most
appliances while using one network.}
\label{tab:per-appliance}
\setlength{\tabcolsep}{4pt}
\begin{tabular*}{\linewidth}{@{\extracolsep{\fill}}ll rrr rrr@{}}
\toprule
& & \multicolumn{3}{c}{\fm (shared)} & \multicolumn{3}{c}{BERT4NILM (per-appliance)} \\
\cmidrule(lr){3-5}\cmidrule(lr){6-8}
Corpus & Appliance & F1 (\%) & AUPRC (\%) & MAE (W) & F1 (\%) & AUPRC (\%) & MAE (W) \\
\midrule
REDD & fridge & 78.7 & 90.4 & 29 & 80.9 & 89.6 & 69 \\
REDD & microwave & 84.4 & 56.0 & 418 & 85.6 & 60.4 & 345 \\
REDD & dishwasher & 53.8 & 77.4 & 60 & 62.6 & 74.4 & 92 \\
\midrule
UKDALE & kettle & 94.6 & 86.7 & 357 & 95.6 & 92.7 & 154 \\
UKDALE & microwave & 79.4 & 62.4 & 402 & 90.1 & 74.5 & 196 \\
UKDALE & fridge & 67.1 & 68.5 & 18 & 71.2 & 76.5 & 34 \\
UKDALE & dishwasher & 39.0 & 88.3 & 78 & 38.9 & 88.7 & 120 \\
UKDALE & toaster & 31.8 & 20.4 & 242 & 45.1 & 27.0 & 386 \\
\midrule
REFIT & microwave & 29.9 & 31.5 & 470 & 13.4 & 28.4 & 1039 \\
REFIT & dishwasher & 14.7 & 53.0 & 256 & 32.4 & 36.5 & 546 \\
REFIT & kettle & 61.5 & 64.5 & 388 & 19.7 & 67.5 & 488 \\
REFIT & washing\_machine & 22.7 & 41.1 & 353 & 42.0 & 36.9 & 434 \\
\bottomrule
\end{tabular*}

\end{table*}

\section{Inference Efficiency}
\label{app:efficiency}

Table~\ref{tab:efficiency} reports single-window latency and parameter counts. \fm
answers any appliance request in 5.5\,ms on four CPU threads and 2.3\,ms on a GPU,
corresponding to roughly 180 CPU requests per second. Its latency is comparable to one
transformer specialist. One 3.7\,M network serves the full request set from a 15\,MB
checkpoint, and each additional requestable appliance adds no model parameters.

\section{Noisy Enrollment}
\label{app:noisy}

Table~\ref{tab:noisy} measures enrollment robustness by corrupting held-out microwave exemplars in three ways. At 40\% per-sample amplitude jitter, the model retains 61\% of clean AUPRC. Truncating 60\% of each activation retains 112\%, consistent with the short discriminative impulse. Label quality has the largest effect: retention is 84\% with 20\% mislabeled exemplars and 12\% with 40\%. Enrollment curation should therefore keep label error below roughly one example in five.

\end{document}